\documentclass{article}
\usepackage{PRIMEarxiv}
\usepackage[utf8]{inputenc} 
\usepackage[T1]{fontenc}    
\usepackage{hyperref}       
\usepackage{url}            
\usepackage{booktabs}       
\usepackage{amsfonts}       
\usepackage{nicefrac}       
\usepackage{microtype}      
\usepackage{lipsum}
\usepackage{fancyhdr}       
\usepackage{graphicx}       
\graphicspath{{media/}}     

\usepackage{natbib}
\usepackage{amsmath} 
\usepackage{amssymb}
\usepackage{xcolor}
\usepackage{subcaption}
\usepackage{tabularx}
\usepackage{multirow}
\usepackage{algorithm}
\usepackage{algpseudocode}
\usepackage{comment}
\usepackage{float}
\usepackage{booktabs}
\usepackage{bm}  
\usepackage{placeins}

\title{CFKS: Continuum-Feature Kernel PDE Solvers}

\author{
  Vikas Dwivedi\thanks{Corresponding Author} \\
  CREATIS Biomedical Imaging Laboratory \\
  INSA, CNRS UMR 5220, Inserm, Universit´e Lyon 1 \\
  Lyon 69621, France\\
  \texttt{vikas.dwivedi@creatis.insa-lyon.fr} \\
   \And
  Balaji Srinivasan \\
  Wadhwani School of Data Science and AI \\
  IIT Madras, India \\
  \texttt{sbalaji@dsai.iitm.ac.in} \\
     \And
  Monica Sigovan \\
  CREATIS Biomedical Imaging Laboratory \\
  INSA, CNRS UMR 5220, Inserm, Universit´e Lyon 1 \\
  Lyon 69621, France\\
  \texttt{monica.sigovan@insa-lyon.fr} \\
    \And
  Bruno Sixou \\
  CREATIS Biomedical Imaging Laboratory \\
  INSA, CNRS UMR 5220, Inserm, Universit´e Lyon 1 \\
  Lyon 69621, France\\
  \texttt{bruno.sixou@insa-lyon.fr} \\
}

\begin{document}
\maketitle

\begin{abstract}
	Neural-network-based PDE solvers require architectural choices,
	particularly the number of layers and neurons per layer.
	Fixing a single Gaussian radial basis function (RBF) feature layer
	and taking the number of features to the continuum limit removes
	the need to prescribe a finite feature count or select individual
	feature centers, thereby reducing representation arbitrariness.
	Motivated by this principle, we introduce Continuum-Feature
	Kernel Solver (CFKS), a deterministic framework for PDEs obtained
	by analytically marginalizing a bounded continuum of Gaussian
	RBF features. CFKS works directly with the resulting
	continuum-feature mean kernel and its operator-applied
	counterparts, parameterized by prescribed Gaussian widths and
	bounded center distributions. For linear PDEs, CFKS leads to
	symmetric operator-kernel collocation systems with null-space
	enforcement of physical constraints. Nonlinear PDEs are addressed
	through linearized initialization and nonlinear least-squares
	residual minimization using analytical kernel derivatives and
	Jacobians. Residual-guided collocation enrichment and sequential
	space--time-slab marching provide additional resolution
	strategies, while prescribed multiscale kernel refinement
	helps resolve localized nonlinear structures.
	Finite-feature experiments confirm convergence of empirical
	value and operator Gram matrices toward their continuum
	counterparts. Numerical studies include oscillatory and
	multiscale ODEs, irregular-domain Poisson and biharmonic problems,
	singular perturbations, transient transport equations, and
	viscous Burgers' equation.
	The results establish CFKS as a direct computational connection
	between finite Gaussian-RBF representations and continuum-feature
	kernel PDE solvers. Codes: \url{https://github.com/vikas-dwivedi-2022/CFKS_Codes}
\end{abstract}


\section{Introduction}
\label{sec:introduction}

Physics-informed neural networks (PINNs) are increasingly used for forward
and inverse partial differential equation (PDE) problems
\citep{RAISSI2019686}. However, their practical performance depends on
architectural choices, particularly the number of layers and neurons
per layer, as well as optimization and collocation-point placement.
A natural way to reduce this architectural arbitrariness is to consider
a single-layer Gaussian radial basis function (RBF) representation
and take the number of features to the continuum limit.
When the corresponding feature expectation is available analytically,
the resulting kernel eliminates the need to prescribe a finite feature
count or select individual feature centers.
The representation nevertheless remains parameterized by prescribed
Gaussian widths and center distributions, which influence approximation
accuracy and numerical conditioning.
This motivates the development of meshfree PDE solvers that work
directly with continuum-feature kernels rather than finite feature
realizations.

A particularly relevant line of development begins with radial-basis-function
(RBF) collocation. Kansa introduced meshfree PDE solvers based on finite RBF
expansions with explicitly prescribed centers
\citep{kansa1990multiquadrics1,kansa1990multiquadrics2}.
Subsequent work provided approximation-theoretic foundations and clarified
limitations of unsymmetric collocation systems
\citep{franke1998solving,hon2001nonsymmetric}.
Symmetric and Hermite formulations subsequently incorporated differential
and boundary operators directly into kernel collocation by applying the
corresponding operators to both arguments of a positive-definite kernel,
thereby constructing symmetric operator-based collocation matrices
\citep{fasshauer2000hermite}.

In parallel, Gaussian-process and neural-network theory established
the correspondence between feature representations and kernels.
Infinite-width neural models may converge to Gaussian processes
\citep{williams1996computing}, while Gaussian-process theory
relates continuous basis-function representations to covariance
kernels \citep{rasmussen2006gaussian}. In such representations,
a deterministic kernel can be constructed by averaging products
of basis functions over a prescribed distribution of feature
parameters. We refer to the resulting kernel as the
\emph{continuum-feature mean kernel}. Neither infinite-feature
limits nor continuum Gaussian-basis representations are new
in themselves. The question addressed in this work is how
to evaluate the bounded Gaussian-RBF mean kernel and its
differential-operator counterparts analytically and employ
them directly in PDE solvers with prescribed physical constraints.

Random-feature methods exploit the converse direction by approximating
this mean kernel using a finite sampled feature map
\citep{rahimi2007random}. This idea has been transferred to PDEs through
random-feature methods that retain a finite randomized trial space
and solve only for output coefficients \citep{chen2022random}.
More recent work has considered localized, high-order, and
operator-aware variants, including residual-Christoffel sampling for
finite operator-applied feature spaces \citep{linghu2026residual}
and analytic partition-of-unity constructions for high-order PDEs
\citep{chen2026analytic}. Kernel and Gaussian-process methods have
likewise been applied directly to differential operators,
probabilistic collocation, and structurally constrained
boundary-value problems
\citep{graepel2003solving,cockayne2017probabilistic,dalton2024boundary}.

The present work connects these developments through
\emph{Continuum-Feature Kernel Solver (CFKS)}, based on the exact
bounded mean kernel induced by a Gaussian RBF feature family.
Unlike finite RBF representations, which require a prescribed
number of features and their individual center locations,
CFKS replaces the finite feature realization with analytical
averaging over a prescribed distribution of centers.
This produces a deterministic continuum-feature mean kernel,
eliminating finite-feature sampling variability while retaining
the Gaussian widths and center distributions as prescribed
parameters.

An important property of this construction is that differential
operators can be applied directly to the mean kernel.
The resulting operator-applied kernels are equivalent to
analytically averaging the products of the corresponding
operator-applied Gaussian features. Consequently, the kernel
matrices required for PDE collocation can be constructed without
explicitly generating finite RBF features. For Gaussian RBFs with
centers uniformly distributed over bounded intervals or boxes,
the mean kernel admits a closed-form expression involving a
squared-exponential factor and error-function terms that account
for the bounded center support
\citep{rasmussen2006gaussian}.

For linear PDEs, CFKS uses the continuum-feature mean kernel and its
operator-applied counterparts to construct symmetric strong-form
collocation systems. Boundary and initial conditions can be enforced
through a null-space parameterization, separating the satisfaction
of physical constraints from the interior PDE approximation.
For nonlinear PDEs, analytical kernel derivatives enable direct
evaluation of the nonlinear residual and its Jacobian. Linearized
solves provide initialization for nonlinear least-squares residual
minimization, while analytical trial-function parameterizations
can enforce the prescribed physical conditions. Problems exhibiting
localized structures can be addressed through multiscale
continuum-feature kernels and residual-guided collocation
enrichment. These extensions are demonstrated using viscous
Burgers' equation with a two-stage predictor--corrector strategy.

\paragraph{Research gap}
Despite substantial developments in RBF collocation, kernel methods,
Gaussian-process formulations, and random-feature PDE solvers,
two methodological gaps motivate the present work:

\begin{itemize}
	\item Finite RBF and random-feature PDE methods generally require
	a prescribed number of features and an explicit realization of
	their centers, whereas conventional kernel collocation typically
	starts from a prescribed kernel. Although continuum-feature
	limits are well established, the direct construction of PDE
	solvers from analytically marginalized bounded Gaussian-RBF
	feature distributions, including the corresponding
	differential-operator kernels, remains comparatively
	underexplored. Such a formulation removes the finite feature
	count and individual center selection while retaining
	prescribed Gaussian widths and center distributions.
	
	\item The enforcement of boundary and initial conditions
	remains an important consideration in meshfree PDE solvers.
	Physical constraints are frequently incorporated through
	penalty terms, weighted least-squares formulations, or
	jointly assembled constraint systems. Although null-space
	methods for enforcing linear constraints are well established
	\citep{ELIADEFALCO2026114553,MISHRA2026119172,MISHRA2026118674},
	their integration with analytically derived continuum-feature
	operator kernels, together with analytical constraint
	embeddings for nonlinear problems, provides a unified approach
	for separating the enforcement of physical conditions from
	the approximation of the governing PDE.
\end{itemize}

\paragraph{Contributions}
The main contributions of this work are:

\begin{itemize}
	\item We develop CFKS by analytically replacing a finite Gaussian-RBF
	feature representation with its exact bounded continuum-feature mean
	kernel. This removes the need to prescribe a finite feature count or
	select individual feature centers, while retaining Gaussian widths
	and center distributions as prescribed parameters.
	
	\item We derive the corresponding differential-operator kernels and
	employ them in symmetric strong-form collocation for linear PDEs.
	Boundary and initial conditions are enforced through a null-space
	parameterization, allowing the governing equations to be solved
	within a constraint-compatible subspace.
	
	\item We extend CFKS to nonlinear PDEs through linearized
	initialization and nonlinear least-squares residual minimization,
	using analytical kernel derivatives and Jacobians. Analytical
	constraint embeddings and multiscale continuum-feature kernels
	further accommodate prescribed physical conditions and localized
	solution structures.
	
	\item We integrate optional residual-adaptive collocation and
	sequential space--time-slab marching into the CFKS framework.
	Residual-guided enrichment concentrates computational resolution
	in regions of under-resolution, while prescribed multiscale
	kernel refinement provides additional flexibility for resolving
	localized structures.
\end{itemize}

Residual-based adaptive sampling is already well established in
meshfree and physics-informed methods
\citep{WU2023115671,LuLu2021}. In CFKS, it is incorporated as an
optional refinement strategy in which independently evaluated PDE
residuals guide the allocation of additional collocation points.
When the mean kernel is fixed, this refinement changes the
collocation and representer sites without altering the underlying
continuum-feature kernel or prescribed physical conditions.
For nonlinear problems, collocation enrichment can additionally
be combined with prescribed changes in kernel scales.
Sequential space--time-slab marching extends this strategy to
time-dependent problems, consistent with earlier time-marching
and causality-aware formulations
\citep{DWIVEDI2021299,WANG2024116813}.

\paragraph{Organization of the paper}
The remainder of this paper is organized as follows.
Section~\ref{sec:method} presents the mathematical formulation of CFKS,
including the continuum-feature mean kernel, operator-based collocation,
constraint enforcement, and extensions to nonlinear PDEs, residual-adaptive
collocation, and sequential space--time slabs.
Section~\ref{sec:test_cases} introduces the numerical benchmark problems.
Section~\ref{sec:results} presents and discusses the numerical results,
including finite-feature convergence and conditioning studies.
Section~\ref{sec:conclusion} summarizes the main findings and outlines
directions for future research.
The appendices provide the connection between finite-feature models
and the CFKS representer formulation, together with the analytical
derivation of the bounded Gaussian continuum-feature kernel.

\section{Mathematical Formulation}
\label{sec:method}

We first develop CFKS for linear differential problems and subsequently
describe its extension to nonlinear PDEs. Let $\Omega$ denote the
computational domain and $\mathbf z$ the vector of independent variables.
For time-dependent problems, $\mathbf z=(\mathbf x,t)$, where
$\mathbf x$ represents the spatial coordinates and $t$ denotes time.
We consider the linear differential problem
\begin{equation}
	\mathcal{L}u(\mathbf z)=f(\mathbf z),
	\qquad
	\mathbf z\in\Omega,
	\label{eq:pde_general}
\end{equation}
where $\mathcal{L}$ is a prescribed linear differential operator,
$u$ is the unknown solution, and $f$ is a known source function.
The problem is supplemented by boundary conditions and, for
time-dependent equations, appropriate initial conditions.

To formulate the numerical problem, the governing PDE and its
physical conditions are expressed through a collection of
linear constraint functionals,
\begin{equation}
	\ell_i[u]=y_i,
	\qquad
	i=1,\ldots,N,
	\label{eq:linear_constraints}
\end{equation}
where $N$ is the total number of discrete constraints,
$\ell_i$ denotes the linear functional associated with
constraint $i$, and $y_i$ is its prescribed value.
For an interior PDE constraint imposed at a collocation point
$\mathbf z_i$, the functional $\ell_i$ applies the differential
operator $\mathcal{L}$ to the solution at that point, with
$y_i=f(\mathbf z_i)$. Boundary and initial conditions are
represented by the corresponding evaluation or differential
functionals at the selected physical constraint locations.

\subsection{From finite RBF features to the continuum-feature mean kernel}
\label{sec:continuum_kernel}

We begin with a finite representation of the unknown solution using
Gaussian radial basis function (RBF) features. Let
$\mathbf z=(z_1,\ldots,z_d)$ denote a point in the computational domain,
where $d$ is the number of independent variables. For time-dependent
problems, these variables include both spatial coordinates and time.

The $m$th Gaussian RBF feature is defined as
\begin{equation}
	\phi_{\theta_m}(\mathbf z)
	=
	\exp\left(
	-\frac{1}{2}
	\sum_{r=1}^{d}
	\frac{(z_r-\theta_{m,r})^2}{\sigma_r^2}
	\right),
	\label{eq:gaussian_rbf_feature}
\end{equation}
where $\theta_m=(\theta_{m,1},\ldots,\theta_{m,d})$ denotes the
center of the $m$th feature, $\theta_{m,r}$ is its center coordinate
in the $r$th direction, and $\sigma_r>0$ is the prescribed Gaussian
width associated with that direction. The index $r$ runs over the
$d$ independent variables. Equal widths in all directions give
the isotropic Gaussian RBF, whereas different widths allow
coordinate-dependent scaling. In space--time problems, separate
spatial and temporal widths can therefore be prescribed.

Using $M$ such features, where $M$ is the total number of Gaussian
RBF features, the finite-feature approximation is expressed as
\begin{equation}
	u_M(\mathbf z)
	=
	\frac{1}{\sqrt{M}}
	\sum_{m=1}^{M}
	\beta_m\phi_{\theta_m}(\mathbf z),
	\label{eq:finite_feature_model}
\end{equation}
where $u_M$ denotes the approximate solution, $m$ indexes the
individual features, and $\beta_m$ is the output coefficient
associated with the $m$th feature. The feature centers and
Gaussian widths are fixed when determining the coefficients
from the governing constraints. The normalization factor
$1/\sqrt{M}$ ensures that the associated feature Gram matrix
has a nontrivial finite limit as the number of features increases.

Recall that the governing PDE and physical conditions are represented
by $N$ linear constraint functionals, denoted by $\ell_i$, with
$i=1,\ldots,N$. Applying these functionals to the finite-feature
representation gives the constraint matrix $Z$, whose entries are
\begin{equation}
	Z_{im}
	=
	\frac{1}{\sqrt{M}}\,
	\ell_i[\phi_{\theta_m}],
	\label{eq:finite_feature_matrix}
\end{equation}
where $Z_{im}$ is the entry associated with constraint $i$
and feature $m$. Thus, $Z$ is an $N\times M$ matrix, and
$\ell_i[\phi_{\theta_m}]$ denotes the result of applying
the $i$th linear constraint functional to the $m$th
Gaussian RBF feature.

The corresponding constraint Gram matrix is obtained by
multiplying $Z$ by its transpose:
\begin{equation}
	(ZZ^\top)_{ij}
	=
	\frac{1}{M}
	\sum_{m=1}^{M}
	\ell_i[\phi_{\theta_m}]
	\ell_j[\phi_{\theta_m}],
	\label{eq:finite_constraint_gram}
\end{equation}
where $Z^\top$ denotes the transpose of $Z$, and
$i,j=1,\ldots,N$ index pairs of constraint functionals.
The matrix $ZZ^\top$ is symmetric and positive semidefinite,
with each entry representing the inner product of two
constraint-applied finite-feature vectors.

To obtain the continuum-feature formulation, suppose that
the Gaussian RBF centers are independently sampled from
a prescribed probability distribution with density
$\rho(\theta)$, where $\theta$ represents a generic
feature-center vector. The distribution specifies the
region and relative weighting of admissible feature centers.
In the present work, bounded uniform center distributions
are used.

Rather than retaining an explicit finite realization of
these centers, we consider the analytical mean of the
products of Gaussian RBF features. This defines the
\emph{continuum-feature mean kernel}
\begin{equation}
	k_{\rho}(\mathbf z,\mathbf z')
	=
	\mathbb{E}_{\theta\sim\rho}
	\left[
	\phi_{\theta}(\mathbf z)
	\phi_{\theta}(\mathbf z')
	\right],
	\label{eq:population_kernel_method}
\end{equation}
where $k_{\rho}$ denotes the mean kernel associated with
the center distribution $\rho$, and $\mathbf z$ and
$\mathbf z'$ are its first and second arguments,
respectively. The notation
$\mathbb{E}_{\theta\sim\rho}$ denotes expectation with
respect to the prescribed probability distribution
of the feature-center vector $\theta$.
The function $\phi_{\theta}$ is the Gaussian RBF
defined in eq. \ref{eq:gaussian_rbf_feature}, with
the generic center $\theta$ replacing the individual
center $\theta_m$.

For independently sampled feature centers, the law of large
numbers implies that, as the number of features $M$ tends
to infinity, each entry of the empirical constraint Gram
matrix converges to its corresponding population value,
provided the relevant feature moments are finite.
For the bounded Gaussian-feature construction considered
here, these conditions are satisfied. Consequently,
\begin{align}
	(ZZ^\top)_{ij}
	&\longrightarrow
	\mathbb{E}_{\theta\sim\rho}
	\left[
	\ell_i[\phi_{\theta}]
	\ell_j[\phi_{\theta}]
	\right]
	\nonumber\\
	&=
	\ell_i^{\mathbf z}
	\ell_j^{\mathbf z'}
	k_{\rho}(\mathbf z,\mathbf z'),
	\label{eq:operator_gram_limit}
\end{align}
where the superscripts on the constraint functionals
identify the kernel argument on which each functional acts.
Specifically, $\ell_i^{\mathbf z}$ acts on the first
argument of the kernel, while $\ell_j^{\mathbf z'}$
acts on the second. Each functional includes the
appropriate differential or evaluation operation
at the location associated with its constraint.

Equation~\ref{eq:operator_gram_limit} establishes that
the continuum-feature constraint Gram matrix is the
analytical expectation of the corresponding finite-feature
Gram matrix. Thus, CFKS replaces the finite feature count
and individual center realization with a deterministic
kernel defined by the prescribed center distribution
and Gaussian widths.

For Gaussian RBF features with centers uniformly distributed
over bounded intervals, the expectation defining the
continuum-feature mean kernel can be evaluated analytically.
In one dimension, the resulting expression consists of
a Gaussian factor multiplied by error-function terms
that account for the finite support of the center
distribution. When the center distribution is a product
of independent uniform distributions over a Cartesian
box, the multidimensional mean kernel is obtained as
the product of the corresponding one-dimensional kernels.
The complete derivation and closed-form expressions
are provided in Appendix~\ref{app:bounded_gaussian_kernel}.

An important property of the continuum-feature mean kernel
is that its derivatives can be obtained directly from
the analytical kernel expression. To describe derivatives
of arbitrary order, let
$\boldsymbol{p}=(p_1,\ldots,p_d)$ and
$\boldsymbol{q}=(q_1,\ldots,q_d)$ denote two multi-indices,
where $p_r$ and $q_r$ are nonnegative integers specifying
the orders of partial differentiation with respect to
the $r$th coordinates of $\mathbf z$ and $\mathbf z'$,
respectively. The operators
$D_{\mathbf z}^{\boldsymbol{p}}$ and
$D_{\mathbf z'}^{\boldsymbol{q}}$ denote the corresponding
mixed partial derivatives with respect to the first
and second kernel arguments.

For the smooth Gaussian features and bounded center
distributions considered here, differentiation can be
interchanged with expectation. Therefore,
\begin{equation}
	D_{\mathbf z}^{\boldsymbol{p}}
	D_{\mathbf z'}^{\boldsymbol{q}}
	k_{\rho}(\mathbf z,\mathbf z')
	=
	\mathbb{E}_{\theta\sim\rho}
	\left[
	D_{\mathbf z}^{\boldsymbol{p}}
	\phi_{\theta}(\mathbf z)
	D_{\mathbf z'}^{\boldsymbol{q}}
	\phi_{\theta}(\mathbf z')
	\right],
	\label{eq:population_derivative_identity}
\end{equation}
which shows that the mixed derivatives of the mean kernel
are equal to the analytical averages of the corresponding
products of differentiated Gaussian RBF features.

In particular, for the prescribed linear differential
operator $\mathcal{L}$ appearing in the governing PDE,
\begin{equation}
	\mathcal{L}_{\mathbf z}
	\mathcal{L}_{\mathbf z'}
	k_{\rho}(\mathbf z,\mathbf z')
	=
	\mathbb{E}_{\theta\sim\rho}
	\left[
	(\mathcal{L}\phi_{\theta})(\mathbf z)
	(\mathcal{L}\phi_{\theta})(\mathbf z')
	\right],
	\label{eq:population_operator_kernel}
\end{equation}
where $\mathcal{L}_{\mathbf z}$ and
$\mathcal{L}_{\mathbf z'}$ indicate application of the
differential operator to the first and second kernel
arguments, respectively. On the right-hand side,
$(\mathcal{L}\phi_{\theta})(\mathbf z)$ denotes the
operator applied to a Gaussian RBF feature and evaluated
at $\mathbf z$, with the analogous interpretation
for the second argument.

Equation~\ref{eq:population_operator_kernel}
establishes that the operator-applied mean kernel
is exactly the population Gram kernel associated
with the corresponding operator-applied Gaussian
RBF features. Hence, the differential-operator kernels
required by CFKS follow directly from the
continuum-feature construction rather than being
introduced independently.

\subsection{Linear CFKS formulation}
\label{sec:kernel_collocation}

We now construct the CFKS approximation for the linear differential
problem introduced in eq. \ref{eq:pde_general}. Recall that
$\ell_i$, with $i=1,\ldots,N$, denotes the linear functional
associated with the $i$th governing PDE, boundary, or initial
constraint, and $N$ is the total number of discrete constraints.
Each functional includes the appropriate differential or evaluation
operation at its prescribed constraint location.

Let $k_{\rho}$ denote the continuum-feature mean kernel defined
in eq. \ref{eq:population_kernel_method}. The CFKS approximation
is expressed in the kernel representer form
\begin{equation}
	u_h(\mathbf z)
	=
	\sum_{j=1}^{N}
	\alpha_j
	\ell_j^{\mathbf z'}
	k_{\rho}(\mathbf z,\mathbf z'),
	\label{eq:kernel_representer}
\end{equation}
where $u_h$ is the kernel approximation to the unknown solution,
$\mathbf z$ is an arbitrary evaluation point,
$\alpha_j$ is the coefficient associated with the $j$th
constraint, and $\mathbf z'$ is the second argument of the
kernel. The superscript on $\ell_j^{\mathbf z'}$ indicates
that the $j$th constraint functional acts on the second
kernel argument, including evaluation at the location
associated with that constraint. When the functional
contains derivatives, these are applied before evaluating
at the corresponding constraint location.

The representer formulation follows from the connection
between finite-feature models and their continuum-kernel
counterparts, as detailed in Appendix~\ref{app:finite_to_kernel}.

Applying each constraint functional $\ell_i$ to
eq. \ref{eq:kernel_representer} produces the linear system
associated with the discrete PDE and physical conditions.
Its coefficient matrix is the operator Gram matrix
$K_{\ell\ell}$, with entries
\begin{equation}
	(K_{\ell\ell})_{ij}
	=
	\ell_i^{\mathbf z}
	\ell_j^{\mathbf z'}
	k_{\rho}(\mathbf z,\mathbf z'),
	\qquad
	i,j=1,\ldots,N,
	\label{eq:kernel_gram}
\end{equation}
where $\ell_i^{\mathbf z}$ and $\ell_j^{\mathbf z'}$
act on the first and second kernel arguments, respectively.
Both functionals include evaluation at their prescribed
constraint locations. Thus, $(K_{\ell\ell})_{ij}$
represents the interaction between the $i$th and
$j$th constraint functionals through the continuum-feature
mean kernel.

The matrix $K_{\ell\ell}$ has dimensions $N\times N$
and is symmetric and positive semidefinite. As established
in eq. \ref{eq:operator_gram_limit}, its entries are
the analytical population limits of the corresponding
finite-feature constraint Gram-matrix entries.

Let
$\boldsymbol{\alpha}=(\alpha_1,\ldots,\alpha_N)^\top$
denote the vector of representer coefficients and
$\mathbf y=(y_1,\ldots,y_N)^\top$ the vector of prescribed
constraint values. A regularized collocation formulation
is then given by
\begin{equation}
	\left(
	K_{\ell\ell}+\lambda I
	\right)
	\boldsymbol{\alpha}
	=
	\mathbf y,
	\label{eq:kernel_coefficients}
\end{equation}
where $I$ is the $N\times N$ identity matrix and
$\lambda\geq0$ is a prescribed regularization parameter.
When $\lambda=0$, the system corresponds to direct
constraint interpolation, provided a solution exists.
A positive regularization parameter improves the
conditioning of the coefficient system but generally
relaxes exact satisfaction of the discrete constraints.
The separate enforcement of linear physical conditions
through a null-space parameterization is described
in the subsequent formulation.

Once the coefficient vector $\boldsymbol{\alpha}$
has been determined, the approximate solution can
be evaluated at any point $\mathbf z$ using
\begin{equation}
	u_h(\mathbf z)
	=
	\mathbf k_{\mathbf z\ell}^{\top}
	\boldsymbol{\alpha},
	\qquad
	(\mathbf k_{\mathbf z\ell})_j
	=
	\ell_j^{\mathbf z'}
	k_{\rho}(\mathbf z,\mathbf z'),
	\label{eq:kernel_prediction}
\end{equation}
where $\mathbf k_{\mathbf z\ell}$ is an $N$-component
column vector containing the kernel interactions
between the evaluation point $\mathbf z$ and the
selected constraint functionals. Its $j$th component
is obtained by applying the $j$th functional to the
second argument of the continuum-feature mean kernel.
The superscript $\top$ denotes vector transposition.

Equations~\ref{eq:population_kernel_method}--%
\ref{eq:kernel_prediction} define the kernel formulation
without retaining the finite number of Gaussian RBF
features or their individual center realizations.
Because the bounded continuum-feature mean kernel
is evaluated analytically, no numerical quadrature
over the feature-center distribution is required.
The resulting PDE approximation is determined by
the prescribed kernel parameters, the selected
constraint locations, and the numerical treatment
of the resulting collocation system.

\subsection{Constraint-compatible enforcement of physical conditions}
\label{sec:constraint_enforcement}

When boundary or initial data are to be satisfied independently of the
PDE least-squares fit, we separate the physical constraints from the
interior PDE equations. Let
\begin{equation}
    A\boldsymbol{\alpha}\approx\mathbf f,
    \qquad
    B\boldsymbol{\alpha}=\mathbf g,
    \label{eq:split_system}
\end{equation}
denote, respectively, the interior PDE equations and the boundary or
initial constraints obtained from the CFKS representation
eq. \ref{eq:kernel_representer}.

A particular coefficient vector satisfying the physical constraints is
chosen as
\begin{equation}
    \boldsymbol{\alpha}_p
    =
    B^{+}\mathbf g,
    \label{eq:particular_solution}
\end{equation}
where $B^{+}$ denotes the Moore--Penrose pseudoinverse. Let the columns
of $Q$ span the numerical null space of $B$,
\begin{equation}
    BQ=0.
    \label{eq:nullspace_basis}
\end{equation}
Every constraint-compatible solution can then be parameterized as
\begin{equation}
    \boldsymbol{\alpha}
    =
    \boldsymbol{\alpha}_p
    +
    Q\boldsymbol{\eta}.
    \label{eq:nullspace_parameterization}
\end{equation}
Substitution into the PDE equations gives the reduced problem
\begin{equation}
    AQ\,\boldsymbol{\eta}
    \approx
    \mathbf f-A\boldsymbol{\alpha}_p.
    \label{eq:reduced_pde_system}
\end{equation}
With Tikhonov regularization,
\begin{equation}
    \boldsymbol{\eta}
    =
    \arg\min_{\boldsymbol{\eta}}
    \left\|
    AQ\boldsymbol{\eta}
    -
    \left(
    \mathbf f-A\boldsymbol{\alpha}_p
    \right)
    \right\|_2^2
    +
    \lambda
    \|\boldsymbol{\eta}\|_2^2.
    \label{eq:nullspace_ridge}
\end{equation}
This construction separates satisfaction of the linear physical
constraints from fitting of the PDE residual and avoids introducing
an arbitrary relative penalty between boundary and interior equations.
In practice, the reduced system is solved using rank-revealing
SVD-based linear algebra.

\subsection{Nonlinear CFKS formulation}
\label{sec:nonlinear_cfks}

We now extend the continuum-feature kernel formulation to nonlinear
differential problems. Consider
\begin{equation}
	\mathcal{N}[u](\mathbf z)=f(\mathbf z),
	\qquad
	\mathbf z\in\Omega,
	\label{eq:nonlinear_pde_general}
\end{equation}
where $\mathcal{N}$ is a prescribed nonlinear differential operator,
$u$ is the unknown solution, $f$ is a known source function, and
$\Omega$ is the computational domain. The operator may depend
nonlinearly on the solution, its derivatives, or both. Appropriate
boundary and, when applicable, initial conditions complete the problem.

Unlike the linear formulation, a nonlinear differential operator
does not generally produce a fixed operator Gram matrix.
Instead, CFKS constructs a deterministic kernel trial
representation and determines its coefficients by minimizing
the nonlinear PDE residual.

Let $k_{\rho}$ be the continuum-feature mean kernel defined in
eq. \ref{eq:population_kernel_method}. We consider the trial
representation
\begin{equation}
	u_h(\mathbf z;\boldsymbol{\alpha})
	=
	u_{\mathrm{base}}(\mathbf z)
	+
	w(\mathbf z)
	\sum_{j=1}^{N_r}
	\alpha_j
	k_{\rho}(\mathbf z,\boldsymbol{\zeta}_j),
	\label{eq:nonlinear_kernel_representer}
\end{equation}
where $u_h$ denotes the approximate solution,
$\boldsymbol{\alpha}=(\alpha_1,\ldots,\alpha_{N_r})^\top$
is the unknown coefficient vector, $N_r$ is the number of
kernel representer sites, and $\boldsymbol{\zeta}_j$ is the
location of the $j$th representer site. The prescribed
functions $u_{\mathrm{base}}$ and $w$ define an additive
component and a multiplicative factor, respectively.

When a suitable analytical constraint embedding is available,
these functions can be selected so that the trial solution
satisfies the prescribed boundary or initial conditions
independently of the coefficient values. An unconstrained
representation is recovered by setting
$u_{\mathrm{base}}=0$ and $w=1$.

The trial representation is linear in
$\boldsymbol{\alpha}$, although the governing PDE residual
is generally nonlinear in these coefficients. Importantly,
$N_r$ determines the dimension of the discrete kernel
representation and does not represent the number of
Gaussian RBF features. Each kernel section is already
obtained by analytically averaging a continuum of
Gaussian RBF features over the prescribed center distribution.

Let $\{\mathbf z_i\}_{i=1}^{N_p}$ denote the interior
PDE collocation locations, where $N_p$ is the number
of discrete PDE constraints. The residual at the
$i$th collocation point is defined as
\begin{equation}
	r_i(\boldsymbol{\alpha})
	=
	\mathcal{N}
	\left[
	u_h(\cdot;\boldsymbol{\alpha})
	\right](\mathbf z_i)
	-
	f(\mathbf z_i),
	\qquad
	i=1,\ldots,N_p,
	\label{eq:nonlinear_residual}
\end{equation}
where the nonlinear operator acts on the trial solution
before evaluation at $\mathbf z_i$.

Collecting the residuals into the vector
$\mathbf r(\boldsymbol{\alpha})
=(r_1,\ldots,r_{N_p})^\top$, the kernel coefficients
are determined through
\begin{equation}
	\boldsymbol{\alpha}^{*}
	=
	\arg\min_{\boldsymbol{\alpha}}
	\frac{1}{2}
	\left\|
	\mathbf r(\boldsymbol{\alpha})
	\right\|_2^2,
	\label{eq:nonlinear_least_squares}
\end{equation}
where $\boldsymbol{\alpha}^{*}$ denotes the optimized
coefficient vector and $\|\cdot\|_2$ is the Euclidean norm.
Thus, the nonlinear PDE is enforced through least-squares
minimization of its discrete residual.

The analytical continuum-feature kernel also permits
direct construction of the residual Jacobian.
Define the coefficient-independent trial functions
\begin{equation}
	\psi_j(\mathbf z)
	=
	w(\mathbf z)
	k_{\rho}(\mathbf z,\boldsymbol{\zeta}_j),
	\qquad
	j=1,\ldots,N_r.
	\label{eq:nonlinear_trial_functions}
\end{equation}
Since the derivative of the trial solution with respect
to $\alpha_j$ is $\psi_j$, the Jacobian entries are
\begin{equation}
	J_{ij}(\boldsymbol{\alpha})
	=
	\frac{\partial r_i}{\partial\alpha_j}
	=
	\left(
	\mathcal{N}'[u_h]\psi_j
	\right)(\mathbf z_i),
	\label{eq:nonlinear_residual_jacobian}
\end{equation}
provided the nonlinear operator is Fréchet differentiable.
Here, $\mathcal{N}'[u_h]$ denotes the linearization of
$\mathcal{N}$ about the current trial solution, acting
on the function $\psi_j$. The Jacobian
$J(\boldsymbol{\alpha})$ has dimensions $N_p\times N_r$.

Because the mean kernel and its derivatives are available
analytically, the residual and Jacobian can be evaluated
without generating finite random-feature realizations.
The nonlinear optimization therefore determines only
the coefficients of the prescribed continuum-feature
kernel representation.

\paragraph{Linearized initialization}

The nonlinear least-squares problem generally requires
an initial coefficient vector. A suitable initialization
can be obtained by linearizing the governing equation
about an available approximation.

Let $u^{(k)}$ denote the current approximation at
iteration $k$, and let $\delta u$ be a correction.
A first-order expansion gives
\begin{equation}
	\mathcal{N}[u^{(k)}+\delta u]
	\approx
	\mathcal{N}[u^{(k)}]
	+
	\mathcal{N}'[u^{(k)}]\delta u,
	\label{eq:nonlinear_linearization}
\end{equation}
where $\mathcal{N}'[u^{(k)}]$ is the Fréchet derivative
of the nonlinear operator evaluated at $u^{(k)}$.
The corresponding linearized equation is
\begin{equation}
	\mathcal{N}'[u^{(k)}]\delta u
	\approx
	f-\mathcal{N}[u^{(k)}].
	\label{eq:nonlinear_increment_equation}
\end{equation}
For a fixed approximation $u^{(k)}$, this is a linear
differential problem for the correction $\delta u$.
Its discretization can use the analytical kernel
derivatives developed for linear CFKS.

Alternatively, a Picard-type approximation may be used
when the nonlinear operator admits an appropriate
coefficient-freezing formulation. The resulting
linearized solution supplies initial coefficients
for the nonlinear least-squares problem
eq. \ref{eq:nonlinear_least_squares}.

This initialization procedure does not replace the
nonlinear coefficient solve. Rather, it provides
a starting approximation for minimizing the full
nonlinear PDE residual.

\paragraph{Multiscale continuum-feature kernels}

For problems with localized gradients or multiple
characteristic scales, the trial representation
can combine continuum-feature mean kernels with
different prescribed Gaussian widths and center
distributions.

We consider an additive global--local kernel
of the form
\begin{equation}
	k_{\mathrm{ms}}^{(s)}(\mathbf z,\mathbf z')
	=
	\widehat{k}_g^{(s)}(\mathbf z,\mathbf z')
	+
	\eta\,
	\widehat{k}_{\ell}^{(s)}(\mathbf z,\mathbf z'),
	\qquad
	\eta>0,
	\label{eq:multiscale_mean_kernel}
\end{equation}
where $k_{\mathrm{ms}}^{(s)}$ is the multiscale kernel
at refinement stage $s$, and $\eta$ is a prescribed
positive mixing weight. The subscripts $g$ and $\ell$
identify the global and local components, respectively.

The global kernel is constructed from a bounded
distribution of Gaussian centers covering the
spatial domain, whereas the local kernel uses
a center distribution concentrated around
a prescribed region of interest. Each component
is an analytically marginalized Gaussian-RBF mean
kernel of the form defined in
eq. \ref{eq:population_kernel_method}.

To balance their contributions, the components
are normalized according to
\begin{equation}
	\widehat{k}_q^{(s)}(\mathbf z,\mathbf z')
	=
	\frac{
		k_q^{(s)}(\mathbf z,\mathbf z')
	}{
		k_q^{(s)}
		(\mathbf z_{\mathrm{ref}},
		\mathbf z_{\mathrm{ref}})
	},
	\qquad
	q\in\{g,\ell\},
	\label{eq:multiscale_kernel_normalization}
\end{equation}
where $q$ identifies the kernel component and
$\mathbf z_{\mathrm{ref}}$ is a prescribed reference
point at which the diagonal kernel value is positive.

Positive normalization factors preserve positive
semidefiniteness, and the positive weighted sum
in eq. \ref{eq:multiscale_mean_kernel} remains
a symmetric positive-semidefinite kernel.
Its derivatives are obtained by differentiating
the analytical component kernels.

The center distributions, Gaussian widths,
normalization point, and mixing weight remain
prescribed parameters. The multiscale construction
therefore increases representation flexibility
without introducing finite Gaussian-feature
center sampling.

\paragraph{Illustration using viscous Burgers' equation}

As a nonlinear example, consider the one-dimensional
viscous Burgers equation
\begin{equation}
	u_t+u\,u_x-\nu u_{xx}=0,
	\qquad
	(x,t)\in[-1,1]\times(0,T],
	\label{eq:burgers}
\end{equation}
where $x$ denotes the spatial coordinate, $t$ is
time, $T$ is the final simulation time, and
$\nu>0$ is the viscosity. The subscripts $t$,
$x$, and $xx$ denote the corresponding first
time, first spatial, and second spatial derivatives.

For the homogeneous Dirichlet boundary conditions
and prescribed initial profile used in this
benchmark, analytical constraint embedding is
applied on successive space--time slabs.

Let $n$ index the time slabs, with
$\tau=t-t_n$ denoting the local time measured
from the beginning $t_n$ of the current slab.
At refinement stage $s$, the trial solution is
\begin{equation}
	u_{h,n}^{(s)}(x,\tau)
	=
	u_{\mathrm{in},n}^{(s)}(x)
	+
	\tau(1-x^2)
	v_{h,n}^{(s)}(x,\tau),
	\label{eq:burgers_embedding}
\end{equation}
where $u_{\mathrm{in},n}^{(s)}$ is the incoming
solution profile and $v_{h,n}^{(s)}$ is an
unknown correction field.

The factors $\tau$ and $1-x^2$ make the
correction vanish at the beginning of each
slab and at the spatial boundaries,
respectively. Hence, the trial solution
preserves the incoming profile and satisfies
the homogeneous boundary conditions, provided
the incoming profile is boundary-compatible.

The correction field is represented by
\begin{equation}
	v_{h,n}^{(s)}(\mathbf z)
	=
	\sum_{j=1}^{N_{p,n}^{(s)}}
	\alpha_{j,n}^{(s)}
	k_{\mathrm{ms}}^{(s)}
	(\mathbf z,\mathbf z_{j,n}^{(s)}),
	\label{eq:burgers_multiscale_representer}
\end{equation}
where $\mathbf z=(x,\tau)$,
$N_{p,n}^{(s)}$ is the number of PDE collocation
points on slab $n$ at stage $s$,
$\mathbf z_{j,n}^{(s)}$ denotes the corresponding
collocation location, and
$\alpha_{j,n}^{(s)}$ is its kernel coefficient.
Thus, the PDE collocation locations also serve
as kernel representer sites.

For a fixed slab and stage, let
$\boldsymbol{\alpha}$ collect the unknown
coefficients, with the slab and stage indices
suppressed for simplicity. Analytical
differentiation of the embedded trial solution
gives the affine relations
\begin{align}
	\mathbf u
	&=
	\mathbf u_{\mathrm{in}}
	+
	D_u\boldsymbol{\alpha},
	\nonumber\\
	\mathbf u_x
	&=
	\mathbf u_{\mathrm{in},x}
	+
	D_{u_x}\boldsymbol{\alpha},
	\nonumber\\
	\mathbf u_{xx}
	&=
	\mathbf u_{\mathrm{in},xx}
	+
	D_{u_{xx}}\boldsymbol{\alpha},
	\nonumber\\
	\mathbf u_t
	&=
	D_{u_t}\boldsymbol{\alpha},
	\label{eq:burgers_affine_derivative_maps}
\end{align}
where the vectors on the left contain the
solution and its derivatives at the PDE
collocation points. The vectors
$\mathbf u_{\mathrm{in}}$,
$\mathbf u_{\mathrm{in},x}$, and
$\mathbf u_{\mathrm{in},xx}$ contain the
incoming profile and its spatial derivatives
at those points.

The matrices $D_u$, $D_{u_x}$,
$D_{u_{xx}}$, and $D_{u_t}$ represent
the corresponding contributions of the
kernel correction, including derivatives
of the analytical embedding factor.
The incoming profile has no local-time
dependence and therefore contributes
nothing to the time derivative.

The discrete nonlinear residual is
\begin{equation}
	\mathbf r(\boldsymbol{\alpha})
	=
	\mathbf u_t
	+
	\mathbf u\odot\mathbf u_x
	-
	\nu\mathbf u_{xx},
	\label{eq:burgers_discrete_residual}
\end{equation}
where $\odot$ denotes componentwise
multiplication. Differentiating this
residual with respect to the kernel
coefficients yields
\begin{equation}
	J(\boldsymbol{\alpha})
	=
	D_{u_t}
	+
	\operatorname{diag}(\mathbf u_x)D_u
	+
	\operatorname{diag}(\mathbf u)D_{u_x}
	-
	\nu D_{u_{xx}},
	\label{eq:burgers_jacobian}
\end{equation}
where $\operatorname{diag}(\cdot)$ denotes
the diagonal matrix formed from a vector.

The nonlinear coefficients are obtained
by minimizing the residual in
eq. \ref{eq:burgers_discrete_residual}
using the analytical Jacobian in
eq. \ref{eq:burgers_jacobian}.
Linearized initialization, multiscale
kernel refinement, and residual-guided
collocation are used in the numerical
implementation, with the specific
settings described in the benchmark
setup.

\subsection{Residual-adaptive collocation}
\label{sec:rac}

The continuum-feature kernel formulation permits either
fixed or adaptive PDE collocation. A fixed deterministic
point distribution is sufficient when it adequately
resolves the solution. For problems with localized
under-resolution, residual-adaptive collocation can
concentrate additional PDE constraints in regions
where the current approximation exhibits larger
governing-equation residuals.

This refinement mechanism is not required by CFKS
itself. It provides an additional resolution strategy
that can be combined with linear or nonlinear kernel
formulations.

To describe both cases, define the PDE residual
operator
\begin{equation}
	\mathcal{R}[u]
	=
	\begin{cases}
		\mathcal{L}u-f,
		& \text{for a linear PDE},
		\\[1mm]
		\mathcal{N}[u]-f,
		& \text{for a nonlinear PDE},
	\end{cases}
	\label{eq:general_residual_operator}
\end{equation}
where $\mathcal{L}$ and $\mathcal{N}$ are the
prescribed linear and nonlinear differential
operators, respectively, and $f$ is the
corresponding source function.

Let $\mathcal{P}^{(0)}$ denote an initial
deterministic PDE collocation set, and let
$u^{(0)}$ be the CFKS approximation obtained
from this set. We refer to $u^{(0)}$ as the
predictor solution.

The computational region is partitioned into
$N_B$ blocks, denoted by
$\{\Omega_b\}_{b=1}^{N_B}$, where $b$ is the
block index. For each block, define an
independent residual probe set
$\mathcal{Q}_b\subset\Omega_b$.
The probe points are distinct from the PDE
collocation locations used to construct the
predictor.

At a probe point $\mathbf q$, the predictor
residual is
\begin{equation}
	r^{(0)}(\mathbf q)
	=
	\mathcal{R}[u^{(0)}](\mathbf q),
	\label{eq:probe_residual}
\end{equation}
where $r^{(0)}$ measures the discrepancy
between the governing differential equation
and the predictor approximation.

The residual magnitude within each block
is characterized by its root-mean-square
and maximum values,
\begin{equation}
	R_b^{\mathrm{rms}}
	=
	\left(
	\frac{1}{|\mathcal Q_b|}
	\sum_{\mathbf q\in\mathcal Q_b}
	|r^{(0)}(\mathbf q)|^2
	\right)^{1/2},
	\qquad
	R_b^{\max}
	=
	\max_{\mathbf q\in\mathcal Q_b}
	|r^{(0)}(\mathbf q)|.
	\label{eq:block_residuals}
\end{equation}
Here, $|\mathcal{Q}_b|$ is the number of
probe points in block $b$,
$R_b^{\mathrm{rms}}$ measures the average
residual magnitude, and $R_b^{\max}$
measures the largest observed residual.

A normalized residual indicator is
constructed by combining these measures:
\begin{equation}
	s_b
	=
	\omega
	\frac{R_b^{\mathrm{rms}}}
	{\max_{1\leq j\leq N_B}
		R_j^{\mathrm{rms}}+\varepsilon}
	+
	(1-\omega)
	\frac{R_b^{\max}}
	{\max_{1\leq j\leq N_B}
		R_j^{\max}+\varepsilon},
	\label{eq:residual_indicator}
\end{equation}
where $s_b$ is the indicator for block $b$,
$\omega\in[0,1]$ is a prescribed weighting
parameter, and $\varepsilon>0$ prevents
division by zero. The index $j$ runs
over all blocks in the region being
refined. The two terms measure,
respectively, the relative average
residual and the relative peak residual.

The indicator determines the refinement
level assigned to each block. For a
three-level strategy, define
\begin{equation}
	\gamma_b=
	\begin{cases}
		\mathrm{high},
		& s_b\geq s_{\mathrm{high}},
		\\[1mm]
		\mathrm{mid},
		& s_{\mathrm{mid}}
		\leq s_b<s_{\mathrm{high}},
		\\[1mm]
		\mathrm{low},
		& s_b<s_{\mathrm{mid}},
	\end{cases}
	\label{eq:adaptation_rule}
\end{equation}
where $\gamma_b$ denotes the refinement
level, and $s_{\mathrm{mid}}$ and
$s_{\mathrm{high}}$ are prescribed
thresholds satisfying
$0\leq s_{\mathrm{mid}}<s_{\mathrm{high}}$.

The levels correspond to prescribed
collocation budgets
$N_{\mathrm{low}}$,
$N_{\mathrm{mid}}$, and
$N_{\mathrm{high}}$.
The number of PDE collocation points
assigned to block $b$, denoted by
$N_b$, is selected according to
its refinement level $\gamma_b$.
The total number of PDE collocation
points after refinement is therefore
\begin{equation}
	N_p^{\mathrm{ref}}
	=
	\sum_{b=1}^{N_B}N_b,
	\label{eq:adaptive_total_points}
\end{equation}
where $N_p^{\mathrm{ref}}$ is the
total refined collocation budget.

Additional implementation rules may
be introduced to avoid abrupt spatial
changes in collocation density, such
as extending high-refinement assignments
to neighboring blocks. The particular
thresholds, point budgets, and
neighboring-block rules are selected
for the numerical problem under study.

Once the blockwise budgets are determined,
the refined PDE collocation locations
are generated within the corresponding
blocks. Deterministic low-discrepancy
sequences provide one means of
constructing these point sets without
introducing random sampling.

For a prescribed continuum-feature
mean kernel, residual-guided
collocation refinement modifies
the discrete PDE constraints but
does not change the analytical
kernel itself. If the kernel
representer sites coincide with
the collocation locations, as
in the nonlinear implementation
considered here, enrichment also
changes the number of kernel
coefficients.

Residual-guided collocation may
additionally be combined with
prescribed changes in Gaussian
kernel widths or center
distributions. In that case,
the discrete collocation
representation and the
underlying kernel parameters
are both modified.

In the viscous Burgers
experiment, residual-adaptive
collocation forms part of the
second stage of a
predictor--corrector procedure.
Residual indicators computed
from the completed predictor
determine the corrector
collocation distribution.
The corrector also employs
narrower prescribed kernel
widths, so the resulting
accuracy improvement cannot
be attributed to collocation
adaptation alone.

The numerical values used
for the residual weights,
refinement thresholds,
probe distributions, and
collocation budgets are
specified with the
corresponding test problem.

\paragraph{Residual-adaptive collocation settings}

The two-stage Burgers computation uses a deterministic
predictor followed by a residual-guided corrector.
The spatial interval is divided into 20 equal blocks.
For each time slab, the predictor uses 10 PDE
collocation points per block, giving 200 points
per slab.

The predictor residual is evaluated independently
on a tensor-product probe grid within each block,
using 11 spatial locations and 7 local-time
locations. Spatial probe positions extend from
5\% to 95\% of the block width, while the
local-time positions range from
$0.08\Delta t$ to $0.96\Delta t$, where
$\Delta t$ is the slab duration.
Thus, each block contains 77 residual probes.

The residual indicator in
eq. \ref{eq:residual_indicator} uses
$\omega=0.7$, assigning weights of
0.7 and 0.3 to the root-mean-square
and maximum residual measures,
respectively. The refinement
thresholds in
eq. \ref{eq:adaptation_rule}
are prescribed as
$s_{\mathrm{mid}}=0.10$
and $s_{\mathrm{high}}=0.35$.

To reduce abrupt changes in
collocation density, high- and
medium-refinement assignments
are extended to immediately
adjacent spatial blocks.
The expansion does not wrap
around the domain boundaries,
and high-refinement assignments
take precedence where the
expanded regions overlap.

The corrector uses the blockwise
collocation budgets
\begin{equation}
	N_{\mathrm{low}}=10,
	\qquad
	N_{\mathrm{mid}}=20,
	\qquad
	N_{\mathrm{high}}=30.
	\label{eq:burgers_adaptive_point_budgets}
\end{equation}
The residual normalization and
refinement decisions are performed
separately for each time slab.
Consequently, the corrector
collocation count varies with
the residual distribution.

Collocation points in both stages
are generated using a deterministic
two-dimensional Halton sequence
with bases 2 and 3. Within each
spatial block, the first $N_b$
Halton points are mapped to the
corresponding spatial and local-time
intervals, where $N_b$ is the
prescribed blockwise point budget.

The Stage~2 corrector combines this
residual-guided collocation
enrichment with narrower
prescribed global and local
Gaussian kernel widths.
The kernel representer sites
coincide with the PDE collocation
locations in both stages.

\subsection{Sequential space--time-slab formulation}
\label{sec:time_marching}

For time-dependent PDEs, CFKS can be applied sequentially over
consecutive space--time slabs rather than constructing a single
kernel approximation over the entire temporal domain. This
strategy reduces the size of the individual space--time
problems and allows the solution to be advanced through a
sequence of local temporal intervals.

Let $\Omega_x$ denote the spatial domain and
$[t_0,T]$ the temporal interval, where $t_0$ is the
initial time and $T$ is the final time.
The temporal interval is partitioned into $N_T$
consecutive slabs,
\begin{equation}
	I_n=[t_n,t_{n+1}],
	\qquad
	n=0,\ldots,N_T-1,
	\label{eq:time_slab_partition}
\end{equation}
where $n$ is the slab index, $N_T$ denotes the
total number of slabs, and the partition points
satisfy
\begin{equation}
	t_0<t_1<\cdots<t_{N_T}=T.
\end{equation}
The duration of slab $n$ is
$\Delta t_n=t_{n+1}-t_n>0$.

Within each slab, the physical time coordinate
is replaced by the local coordinate
\begin{equation}
	\tau=t-t_n,
	\qquad
	0\leq\tau\leq\Delta t_n,
	\label{eq:local_slab_time}
\end{equation}
where $\tau=0$ and $\tau=\Delta t_n$
correspond to the beginning and end
of the current slab, respectively.

The CFKS approximation on slab $n$ is denoted
by $u_n(\mathbf x,\tau)$, where $\mathbf x$
represents the spatial coordinates.
The governing PDE is enforced at selected
collocation points within the corresponding
space--time region
$\Omega_x\times(0,\Delta t_n]$.
The same continuum-feature kernel formulation
developed in the preceding sections applies
to each slab.

The local coordinate system permits the same
analytical mean-kernel construction to be used
across different slabs. When the slab durations,
local collocation configurations, and prescribed
kernel parameters are identical, the corresponding
kernel matrices and analytical derivative
evaluations may also be reused.

\paragraph{Sequential transfer of the solution}

The initial condition for each slab is obtained
from the prescribed initial data or from the
terminal solution of the preceding slab.

Let $u_{\mathrm{init}}(\mathbf x)$ denote the
prescribed initial profile at time $t_0$.
The sequential transfer condition is
\begin{equation}
	u_n(\mathbf x,0)
	=
	\begin{cases}
		u_{\mathrm{init}}(\mathbf x),
		& n=0,
		\\[1mm]
		u_{n-1}(\mathbf x,\Delta t_{n-1}),
		& n>0,
	\end{cases}
	\label{eq:slab_marching}
\end{equation}
where $u_{n-1}(\mathbf x,\Delta t_{n-1})$
represents the terminal profile of the
preceding slab.

Thus, each local space--time problem receives
its initial data from the preceding time
interval. The solution is advanced
sequentially until the final time is reached.

The incoming profile may be imposed through
discrete initial constraints or incorporated
analytically into a constraint-compatible
trial representation. In either case, the
local approximation is constructed to match
the prescribed incoming data according to
the chosen constraint-enforcement procedure.

In numerical implementations, the terminal
profile may be transferred directly or
represented through an interpolation
procedure. When interpolation is used,
the incoming profile is an approximation
of the preceding terminal solution, and
the transfer accuracy becomes part of
the overall temporal discretization error.

Within each slab, the CFKS representation
remains global over the spatial domain.
The temporal decomposition therefore does
not introduce artificial spatial interfaces
or require spatial domain decomposition.

\paragraph{Predictor--corrector slab marching}

Sequential slab marching can also be combined
with residual-adaptive collocation.
In this case, a predictor solution is first
computed using an initial deterministic
collocation distribution.

Let $u_n^{(0)}(\mathbf x,\tau)$ denote the
predictor approximation on slab $n$,
where the superscript $(0)$ identifies
the predictor stage. The predictor
satisfies the sequential transfer condition
\begin{equation}
	u_n^{(0)}(\mathbf x,0)
	=
	\begin{cases}
		u_{\mathrm{init}}(\mathbf x),
		& n=0,
		\\[1mm]
		u_{n-1}^{(0)}
		(\mathbf x,\Delta t_{n-1}),
		& n>0.
	\end{cases}
	\label{eq:predictor_march}
\end{equation}
The predictor is advanced through the
entire temporal interval, with each slab
initialized from the preceding predictor
slab.

After completing the predictor march,
the PDE residual is evaluated at
independent probe locations, as described
in Section~\ref{sec:rac}.
The resulting residual indicators identify
regions requiring additional collocation
resolution.

Let $\gamma_{b,n}$ denote the refinement
level assigned to block $b$ on slab $n$,
where $b$ is the spatial-block index.
These refinement assignments determine
the collocation distribution used for
the subsequent corrector computation.

The corrector solution, denoted by
$u_n^{(1)}(\mathbf x,\tau)$, is then
computed through a second sequential
march, with
\begin{equation}
	u_n^{(1)}(\mathbf x,0)
	=
	\begin{cases}
		u_{\mathrm{init}}(\mathbf x),
		& n=0,
		\\[1mm]
		u_{n-1}^{(1)}
		(\mathbf x,\Delta t_{n-1}),
		& n>0.
	\end{cases}
	\label{eq:corrector_march}
\end{equation}
where the superscript $(1)$ identifies
the corrector stage.

The predictor and corrector marches are
performed independently, each beginning
from the prescribed initial condition.
The predictor residual information
determines the refined collocation
distributions, but the predictor
solution is not imposed as the
initial profile of the corrector.

Instead, continuity between successive
corrector slabs is maintained through
the terminal profile of the preceding
corrector slab, subject to the selected
numerical transfer procedure.

The corrector may retain the prescribed
mean kernel used by the predictor or
employ different prescribed kernel
parameters. In the latter case,
residual-guided collocation refinement
and kernel-scale refinement are
combined within the same sequential
solution strategy.

\paragraph{General characteristics}

Sequential space--time-slab marching is
compatible with both linear and nonlinear
CFKS formulations. The governing PDE
is discretized independently within
each slab, while the incoming solution
profile supplies the initial data
required for temporal advancement.

For linear problems, the local
coefficients can be determined through
the corresponding kernel collocation
or constraint-compatible linear
least-squares formulation.
For nonlinear problems, the same
slab decomposition can be combined
with linearized initialization and
nonlinear residual minimization.

The continuum-feature mean kernel
and its analytical differential
operators remain applicable within
every slab. Temporal decomposition
changes the organization of the
discrete PDE solve without
reintroducing finite Gaussian
RBF feature realizations.

Residual-guided predictor--corrector
marching is an additional refinement
strategy rather than a requirement
of the sequential formulation.

\section{Test Cases}
\label{sec:test_cases}

We consider a set of ODE and PDE benchmarks spanning oscillatory and
multiscale one-dimensional problems, irregular two-dimensional geometries,
a fourth-order equation, a singularly perturbed boundary layer,
time-dependent transport, and a nonlinear viscous conservation law.
The governing equations, computational domains, physical conditions, and
reference solutions are summarized in Table~\ref{tab:test_cases}.
For the Poisson and biharmonic problems, the five-petal domain is centered at
$(0.5,0.5)$ and its boundary is defined by
$r(\theta)=0.33+0.10\cos(5\theta)$ for $0\leq\theta<2\pi$.

\begin{table*}[t]
	\centering
	\caption{Summary of the numerical test problems.}
	\label{tab:test_cases}
	\footnotesize
	\setlength{\tabcolsep}{4pt}
	\renewcommand{\arraystretch}{1.15}
	
	\begin{tabularx}{\textwidth}{
			p{2.25cm}
			p{3.4cm}
			X
			X}
		\hline
		\textbf{Problem}
		&
		\textbf{Governing equation and domain}
		&
		\textbf{Conditions}
		&
		\textbf{Reference solution}
		\\
		\hline
		
		First-order ODE
		&
		$u'=\cos(15x)$, \quad $x\in[-2\pi,2\pi]$
		&
		$u(0)=0$
		&
		$u_{\mathrm{ex}}=\frac{1}{15}\sin(15x)$
		\\[1mm]
		
		Second-order ODE
		&
		$u''=\sin(15x)$, \quad $x\in[-2\pi,2\pi]$
		&
		$u(0)=0$, \quad $u'(0)=-1/15$
		&
		$u_{\mathrm{ex}}=-\frac{1}{225}\sin(15x)$
		\\[1mm]
		
		Multiscale ODE
		&
		$u'=\cos x+15\cos(15x)$, \quad $x\in[-2\pi,2\pi]$
		&
		$u(0)=0$
		&
		$u_{\mathrm{ex}}=\sin x+\sin(15x)$
		\\[1mm]
		
		Poisson
		&
		$-\Delta u=f$ on the five-petal domain
		&
		$u=u_{\mathrm{ex}}$ on $\partial\Omega$
		&
		$u_{\mathrm{ex}}=\sin(4\pi x)\cos(4\pi y)$; 
		$f=32\pi^2u_{\mathrm{ex}}$
		\\[2mm]
		
		Biharmonic
		&
		$\Delta^2u=f$ on the five-petal domain
		&
		$u=u_{\mathrm{ex}}$, \quad
		$\partial_nu=\nabla u_{\mathrm{ex}}\!\cdot\!\mathbf n$
		on $\partial\Omega$
		&
		$u_{\mathrm{ex}}=\sin(5\pi x)\cos(5\pi y)$;
		$f=2500\pi^4u_{\mathrm{ex}}$
		\\[2mm]
		
		Singular perturbation
		&
		$u'-\nu u''=0$, \quad
		$x\in(0,1)$, $\nu=10^{-3}$
		&
		$u(0)=0$, \quad $u(1)=1$
		&
		$\displaystyle
		u_{\mathrm{ex}}=
		\frac{e^{(x-1)/\nu}-e^{-1/\nu}}
		{1-e^{-1/\nu}}$
		\\[3mm]
		
		Linear advection
		&
		$u_t+u_x=0$, \quad
		$(x,t)\in[-1,1]\times(0,1]$
		&
		$u(-1,t)=u(1,t)$; \quad
		$u(x,0)=e^{-(25x)^2}$
		&
		$\displaystyle
		u_{\mathrm{ex}}
		=
		\exp\!\left[
		-25^2\left(
		\operatorname{mod}(x-t+1,2)-1
		\right)^2
		\right]$
		\\[3mm]
		
		Advection--diffusion
		&
		$u_t+a u_x-\nu u_{xx}=0$, \quad
		$(x,t)\in[0,1]\times(0,1]$,
		$a=0.5$, $\nu=10^{-3}$
		&
		$u(x,0)=e^{-(x-0.2)^2/\nu}$; \newline
		$u(0,t)=u_{\mathrm{ex}}(0,t)$,
		$u(1,t)=u_{\mathrm{ex}}(1,t)$
		&
		$\displaystyle
		u_{\mathrm{ex}}
		=
		\frac{1}{\sqrt{4t+1}}
		\exp\!\left[
		-\frac{(x-0.2-at)^2}
		{\nu(4t+1)}
		\right]$
		\\[3mm]
		
		Viscous Burgers
		&
		$u_t+u\,u_x-\nu u_{xx}=0$, \quad
		$(x,t)\in[-1,1]\times(0,1]$,
		$\nu=0.01$
		&
		$u(x,0)=-\sin(\pi x)$; \newline
		$u(-1,t)=u(1,t)=0$
		&
		Cole--Hopf reference solution
		\\
		
		\hline
	\end{tabularx}
\end{table*}
\section{Results and Discussion}
\label{sec:results}

\subsection{ODE benchmarks}
\label{sec:ode_results}

\begin{figure*}[t]
	\centering
	
	\includegraphics[width=\textwidth]{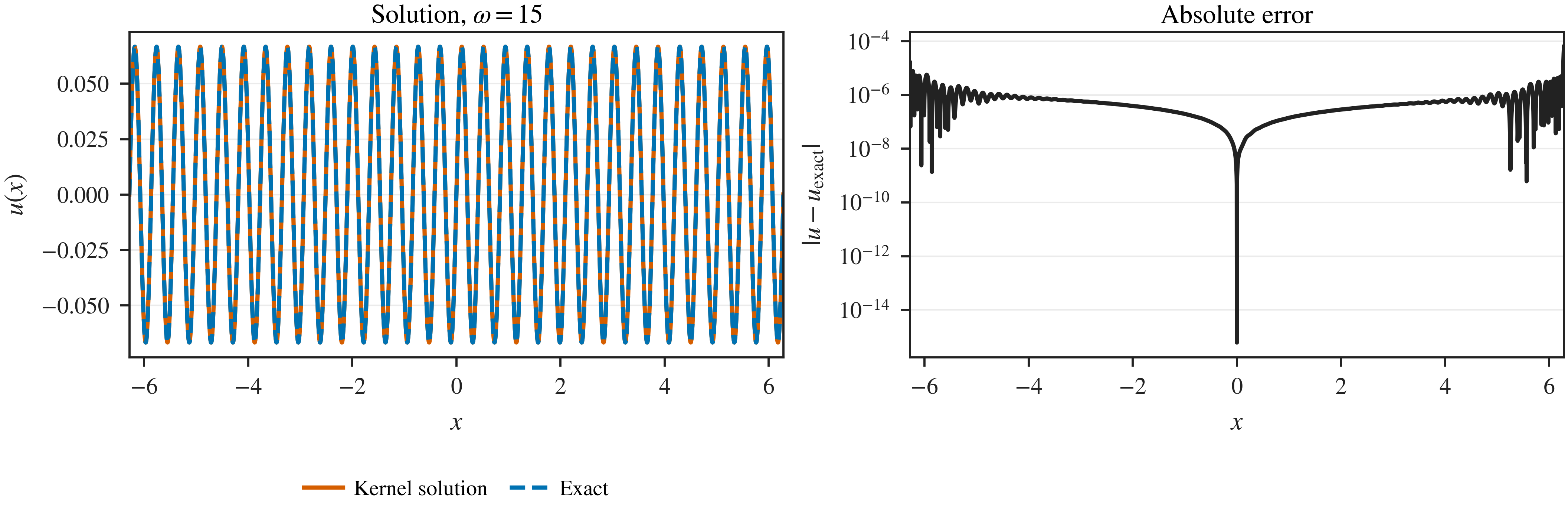}
	
	\vspace{0.6em}
	
	\includegraphics[width=\textwidth]{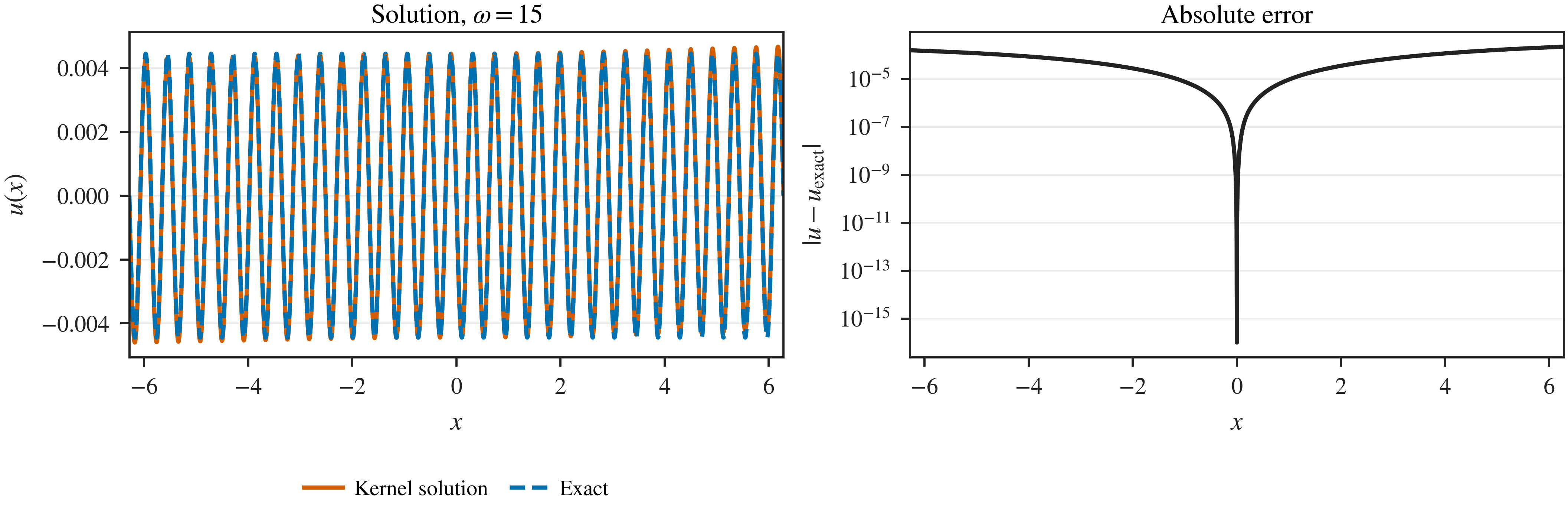}
	
	\vspace{0.6em}
	
	\includegraphics[width=\textwidth]{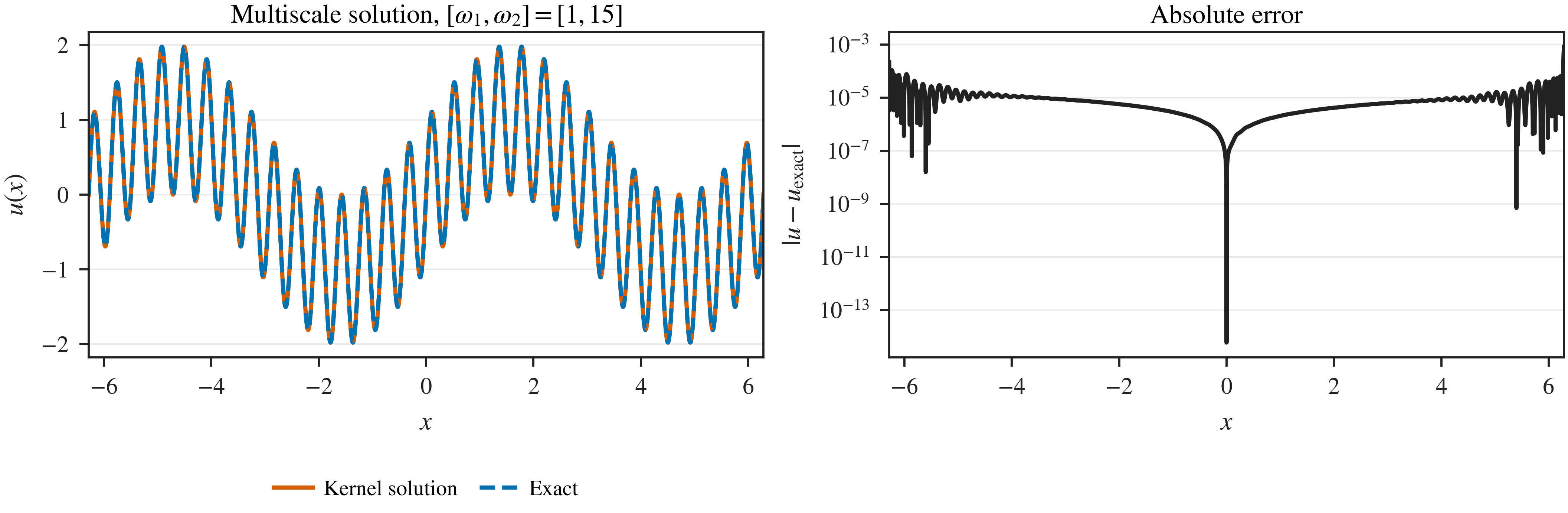}
	
	\caption{ODE benchmarks. Top: first-order oscillatory problem
		$u'=\cos(15x)$. Middle: second-order oscillatory problem
		$u''=\sin(15x)$. Bottom: multiscale problem with
		$[\omega_1,\omega_2]=[1,15]$. In each row, the left panel compares the
		continuum-feature kernel and exact solutions, while the right panel shows
		the pointwise absolute error.}
	\label{fig:ode_results}
\end{figure*}

Figure~\ref{fig:ode_results} summarizes the three one-dimensional ODE
benchmarks. In all cases, the continuum-feature kernel solution closely
tracks the exact solution over the full interval. For the first-order
oscillatory problem, the solution RMSE and maximum error are
$1.70\times10^{-6}$ and $6.30\times10^{-5}$, respectively. The
second-order problem gives an RMSE of $9.96\times10^{-5}$ and a maximum
error of $2.24\times10^{-4}$, while the multiscale case yields
$2.24\times10^{-5}$ and $8.28\times10^{-4}$.

The corresponding residual RMS values are $1.74\times10^{-4}$,
$2.03\times10^{-4}$, and $2.26\times10^{-3}$ for the first-order,
second-order, and multiscale problems, respectively. In each case, the
pointwise error is smallest at the constrained location $x=0$ and tends to
increase toward the ends of the interval. The imposed value constraint is
satisfied to machine precision in all three cases; for the second-order
problem, the derivative condition $u'(0)=-1/15$ is likewise recovered to
machine precision.

Overall, these results verify the first- and second-derivative population
kernels and show that the same continuum-feature representation can resolve
both rapidly oscillating and multiscale solutions without introducing
separate finite feature sets.

\begin{table}[t]
	\centering
	\caption{Computational timing for the three ODE benchmarks.}
	\label{tab:ode_timing}
	\small
	\begin{tabular}{lcccc}
		\hline
		\textbf{Case}
		&
		\textbf{Kernel assembly}
		&
		\textbf{Null-space}
		&
		\textbf{Reduced SVD/solve}
		&
		\textbf{Total solve}
		\\
		&
		\textbf{(s)}
		&
		\textbf{(s)}
		&
		\textbf{(s)}
		&
		\textbf{(s)}
		\\
		\hline
		First-order ODE
		& 1.518 & 0.022 & 0.603 & 2.155 \\
		Second-order ODE
		& 2.427 & 0.036 & 0.823 & 3.296 \\
		Multiscale ODE
		& 1.778 & 0.004 & 0.759 & 2.546 \\
		\hline
	\end{tabular}
\end{table}
Table~\ref{tab:ode_timing} reports the computational cost of the principal
stages of the continuum-feature solve. For all three problems, the dominant
cost is the assembly of the operator-kernel matrices, followed by the SVD of
the reduced system. The null-space construction contributes only a small
fraction of the total cost. The second-order problem is moderately more
expensive because additional differentiated kernel blocks are required to
represent both the second-order differential operator and the two physical
constraints.
\subsection{Comparison with PINNs and FBPINNs}
\label{sec:ode_comparison}

To contextualize the performance of CFKS, we compare it with standard
physics-informed neural networks (PINNs) and finite-basis PINNs (FBPINNs).
The three ODE benchmarks considered above coincide with those studied by
Moseley et al.~\citep{Moseley2023}, allowing comparison under the same
operator families and frequency content.

Standard PINNs exhibit pronounced spectral bias on these problems,
particularly for the $\omega=15$ and multiscale cases. FBPINNs improve the
high-frequency representation by decomposing the domain into overlapping
subdomains with locally normalized subnetworks, but require the selection of
the subdomain layout, overlap, network architecture, and training schedule
\citep{Moseley2023}. For the first-order oscillatory problem, the reported
FBPINN configuration uses 30 subdomains and 50,000 Adam iterations, whereas
the second-order problem requires a specialized sequential
``learning-outwards'' schedule extending to approximately $5\times10^5$
iterations \citep{Moseley2023}.

CFKS does not employ gradient-based training or an explicit finite feature
realization. The three problems are solved through deterministic
population-kernel assembly followed by the null-space and reduced SVD
solves. The measured total assembly-and-solve times are approximately
$2.15$, $3.30$, and $2.55$~s for the first-order, second-order, and
multiscale problems, respectively. The corresponding solution RMSEs are
$1.70\times10^{-6}$, $9.96\times10^{-5}$, and $2.24\times10^{-5}$.
Table~\ref{tab:pinn_fbpinn_cfks} summarizes the comparison.

\begin{table*}[t]
	\centering
	\caption{Qualitative and quantitative comparison of PINN, FBPINN, and CFKS
		on the oscillatory and multiscale one-dimensional ODEs. PINN and FBPINN
		results are taken from \citet{Moseley2023}.}
	\label{tab:pinn_fbpinn_cfks}
	\small
	\begin{tabularx}{\textwidth}{p{2.0cm} X X X}
		\hline
		\textbf{Method}
		&
		\textbf{Training / solution cost}
		&
		\textbf{Architectural complexity}
		&
		\textbf{Accuracy on high-frequency and multiscale tests}
		\\
		\hline
		
		PINN
		&
		Approximately $5\times10^4$--$10^5$ gradient steps; convergence deteriorates
		for high-frequency problems
		&
		Single neural network; performance depends on depth, width, activation
		choice, and may benefit from Fourier features
		&
		Large errors for $\omega=15$ relative to FBPINNs; affected by spectral bias
		\citep{Moseley2023}
		\\[2mm]
		
		FBPINN
		&
		Approximately $5\times10^4$--$5\times10^5$ gradient steps depending on the
		problem
		&
		Typically 20--30 overlapping subdomains with multiple local networks;
		requires subdomain, overlap, and training-schedule choices
		&
		Accurately resolves the high-frequency and multiscale solutions; best
		reported first-order errors are of order $10^{-4}$
		\citep{Moseley2023}
		\\[2mm]
		
		CFKS
		&
		No backpropagation; deterministic kernel assembly and linear-algebra solve.
		Total solve times: $2.15$--$3.30$~s
		&
		No neural architecture, finite hidden-layer width, or sampled center
		realization; prescribed kernel parameters and collocation constraints
		&
		Solution RMSEs of $1.70\times10^{-6}$, $9.96\times10^{-5}$, and
		$2.24\times10^{-5}$ for the first-order, second-order, and multiscale
		problems, respectively
		\\
		\hline
	\end{tabularx}
\end{table*}

The comparison highlights a different computational route to high-frequency
resolution. FBPINNs address spectral bias by decomposing the domain and
training multiple local neural networks, whereas CFKS works directly with
the deterministic population kernel associated with the Gaussian feature
family. Consequently, the three ODEs require no backpropagation, neural
architecture selection, or finite-center sampling. The present timings
should not be interpreted as a direct hardware-normalized speed comparison
with the results of \citep{Moseley2023}; rather, they show that the
population-kernel formulation attains comparable high-frequency resolution
through a small number of dense linear-algebra operations.

\subsection{Poisson and biharmonic problems on the five-petal domain}
\label{sec:flower_results}

\begin{figure*}[t]
	\centering
	
	\includegraphics[width=\textwidth]{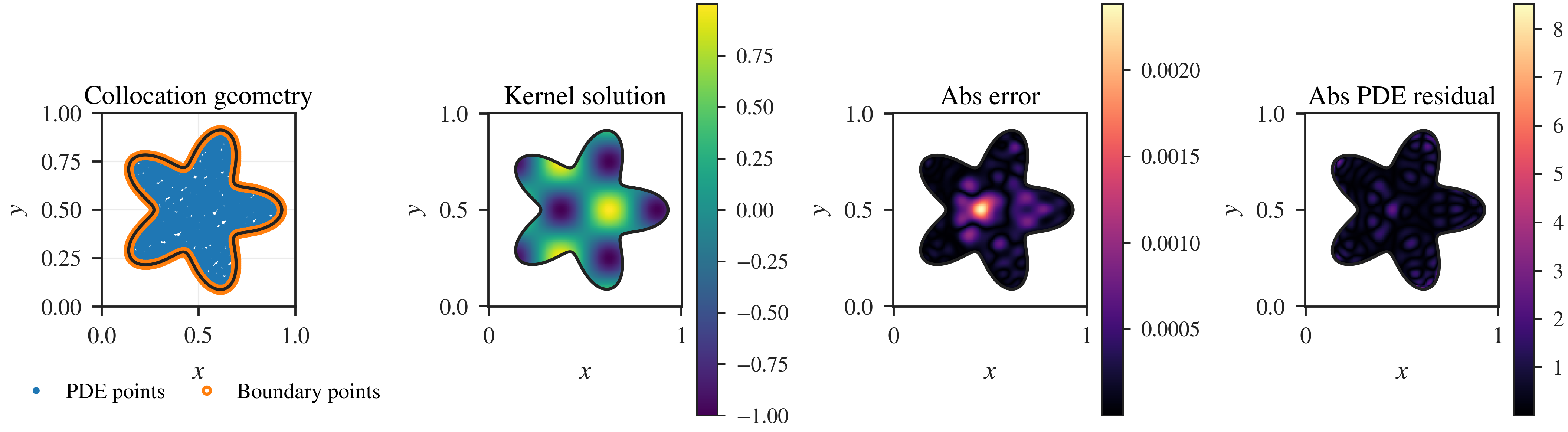}
	
	\vspace{0.8em}
	
	\includegraphics[width=\textwidth]{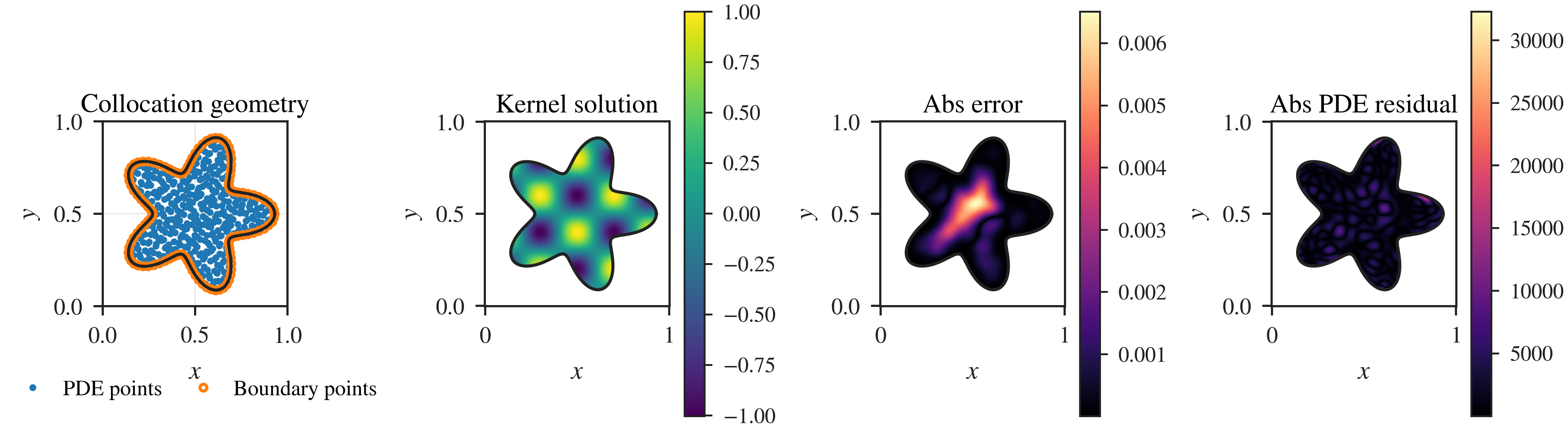}
	
	\caption{CFKS results on the five-petal domain. Top row: Poisson
		problem. Bottom row: biharmonic problem. In each row, the panels show
		the collocation geometry, CFKS solution, pointwise absolute error, and
		absolute PDE residual.}
	\label{fig:flower_pdes}
\end{figure*}

Figure~\ref{fig:flower_pdes} summarizes the CFKS results for the Poisson
and biharmonic problems on the irregular five-petal domain. The top row
corresponds to the second-order Poisson problem and the bottom row to the
fourth-order biharmonic problem.

For the Poisson case, CFKS reproduces the manufactured solution over the
entire nonconvex domain with an independent solution RMSE of
$4.32\times10^{-4}$ and a maximum absolute error of
$2.37\times10^{-3}$. The maximum boundary error is
$2.23\times10^{-5}$, confirming that the Dirichlet constraint is maintained
accurately by the null-space formulation. The independent PDE residual has
RMS $7.32\times10^{-1}$ and maximum magnitude $8.79$. As seen in the top
row of Fig.~\ref{fig:flower_pdes}, the largest solution errors remain
localized within the interior rather than being distributed uniformly over
the domain.

The biharmonic problem provides a more demanding test because it requires
fourth-order population-kernel derivatives together with simultaneous
Dirichlet and normal-derivative boundary conditions. CFKS achieves an
independent solution RMSE of $1.71\times10^{-3}$ and a maximum absolute
error of $6.51\times10^{-3}$. The maximum training boundary error is
$1.21\times10^{-5}$, while the dense Dirichlet-boundary error is
$1.05\times10^{-6}$. The biharmonic residual has RMS
$2.83\times10^{3}$ and maximum magnitude $3.09\times10^{4}$. Although the
absolute residual is larger than in the Poisson case, this reflects the
substantially larger scale introduced by repeated differentiation in the
fourth-order operator; the solution itself remains accurate throughout the
domain. The total CFKS solve time is $0.736$~s, comprising approximately
$0.663$~s for kernel/operator assembly, $0.046$~s for the null-space
construction, and $0.020$~s for the reduced SVD solve. The dominant
computational cost is therefore the assembly of the high-order
operator-kernel matrices rather than the constrained linear solve.

The two cases therefore show that the same CFKS construction extends from
second- to fourth-order operators on irregular geometries without altering
the underlying population-kernel representation. They also demonstrate
that the center-support geometry need not coincide with the physical
domain: the nonconvex boundary enters through the physical constraint
functionals, while the population kernel is defined on the surrounding
embedding region.

\subsection{Singularly perturbed problem}
\label{sec:spp_results}

\begin{figure*}[t]
	\centering
	\includegraphics[width=\textwidth]{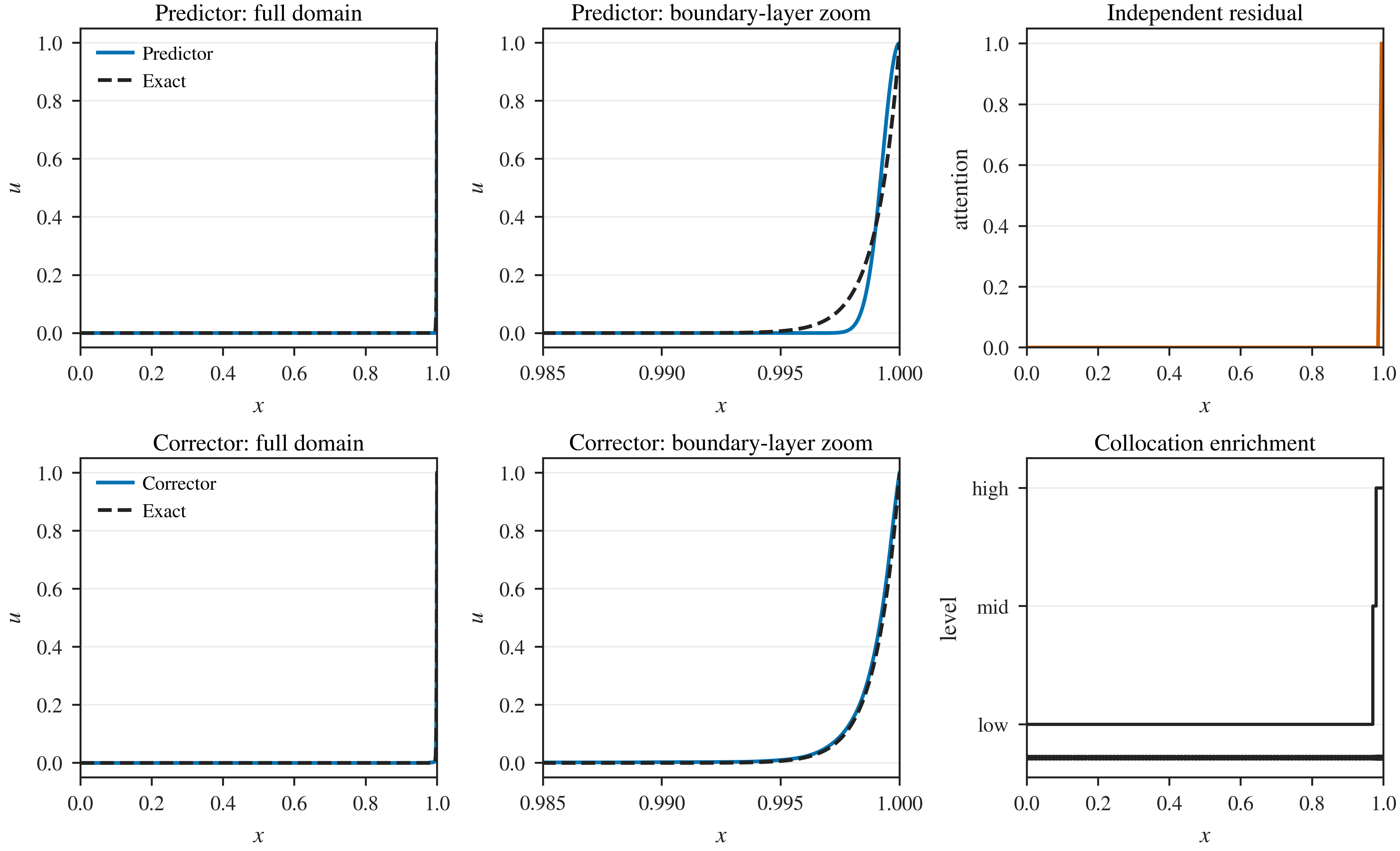}
	\caption{CFKS solution of the singularly perturbed problem. Top row:
		coarse predictor over the full domain, boundary-layer zoom, and
		independent residual indicator. Bottom row: adaptive corrector over the
		full domain, boundary-layer zoom, and residual-guided collocation
		enrichment. The adaptive refinement is concentrated near the boundary
		layer at $x=1$.}
	\label{fig:spp_results}
\end{figure*}

Figure~\ref{fig:spp_results} illustrates the predictor--residual--corrector
behavior for the singularly perturbed boundary-value problem with
$\nu=10^{-3}$. The coarse predictor captures the outer solution but
under-resolves the narrow boundary layer near $x=1$. This is reflected by
the independent residual indicator, which remains negligible over most of
the interval and rises sharply only in the boundary-layer region. As a
result, the adaptive strategy assigns 97 blocks to the low level, one to the
mid level, and only two to the high level.

The predictor uses 200 PDE collocation points and yields an RMSE of
$5.95\times10^{-3}$ with a maximum error of $1.82\times10^{-1}$. After
residual-guided enrichment, the corrector uses 250 PDE points and reduces
the RMSE to $1.76\times10^{-3}$ and the maximum error to
$6.50\times10^{-2}$. Thus, the adaptive corrector improves the RMSE by a
factor of $3.38$ and the maximum error by a factor of $2.80$ relative to a
budget-matched uniform discretization using the same 250 PDE points. The
uniform reference remains essentially unchanged from the coarse predictor,
with an RMSE of $5.95\times10^{-3}$, showing that the improvement is due to
where the additional constraints are placed rather than simply to their
number.

The adaptive set uses only $10.42\%$ of the fully enriched budget obtained
by assigning 24 PDE points to every block. The predictor kernel assembly and
spectral solve require approximately $0.047$ and $0.030$~s, respectively,
while residual probing and level assignment require about $0.191$~s in the
recorded run. The adaptive corrector requires approximately $0.090$~s for
kernel assembly and $0.034$~s for the spectral solve, giving a total CFKS
corrector solve time of $0.123$~s. These results show that the
residual-adaptive stage concentrates additional constraints in the
boundary-layer region while leaving the underlying CFKS representation
unchanged.

For context, the same singularly perturbed steady
advection--diffusion problem with $\nu=10^{-3}$ was considered by
De Florio et al.~\citep{DEFLORIO2024115396}. Their experiments used
$N_{\mathrm{int}}=10^4$ interior collocation points for all methods,
while the accurate ELM-based PIELM and X-TFC configurations also employed
$10^4$ hidden neurons at this parameter value. In contrast, the present
CFKS predictor uses only 200 PDE collocation points, and the
residual-adaptive corrector increases this number to 250 by concentrating
additional points near the localized boundary layer. Thus, rather than
increasing the global representation and collocation budgets, CFKS exploits
the independently evaluated residual to allocate resolution only where it is
needed.

\subsection{Advection and advection--diffusion}
\label{sec:transport_results}

\begin{figure*}[t]
	\centering
	\includegraphics[width=\textwidth]{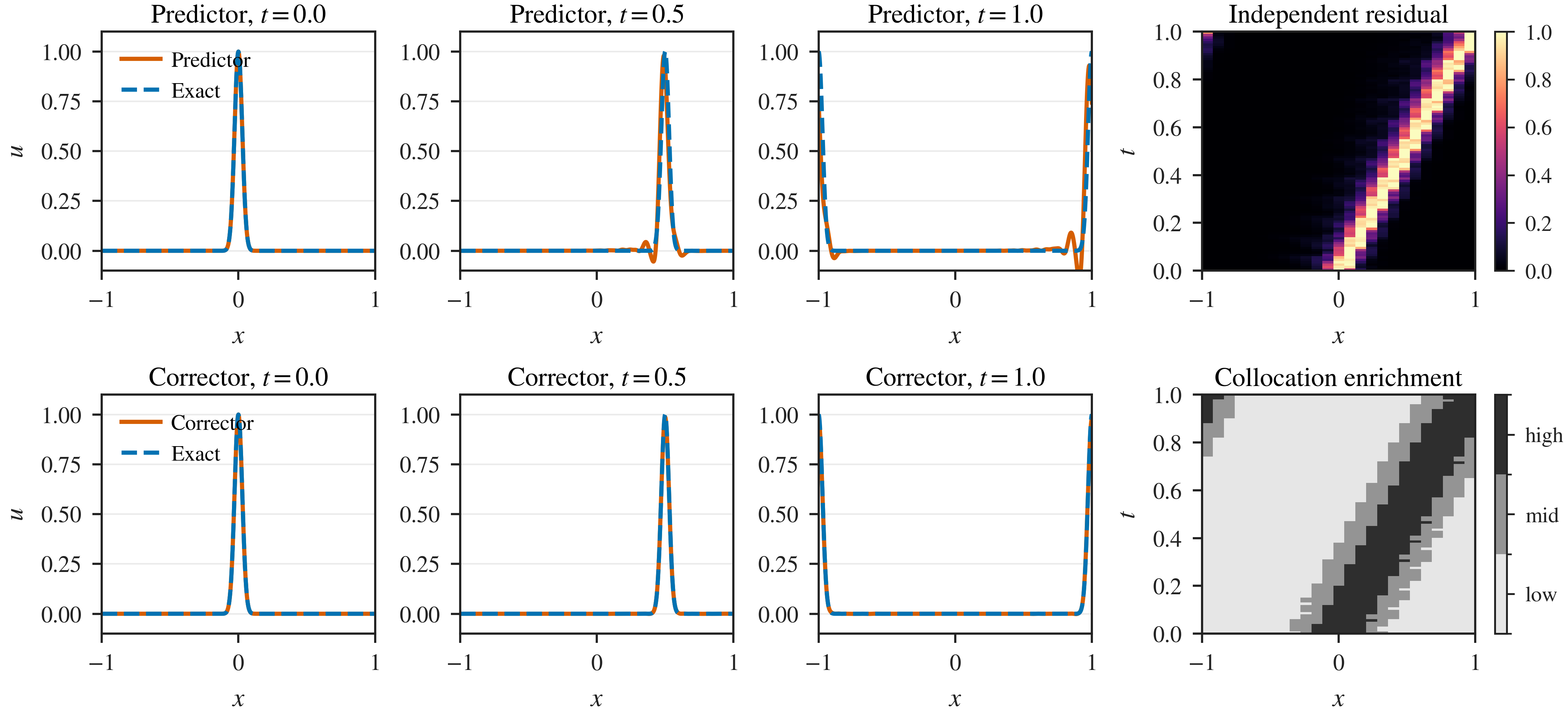}
	\caption{CFKS results for the linear advection problem.
		Top row: predictor solutions at representative times and
		the independently evaluated PDE residual. Bottom row:
		adaptive corrector solutions and the resulting space--time
		collocation enrichment. The enriched collocation region
		follows the trajectory of the advecting Gaussian pulse.}
	\label{fig:advection_results}
\end{figure*}

\begin{figure*}[t]
	\centering
	\includegraphics[width=\textwidth]{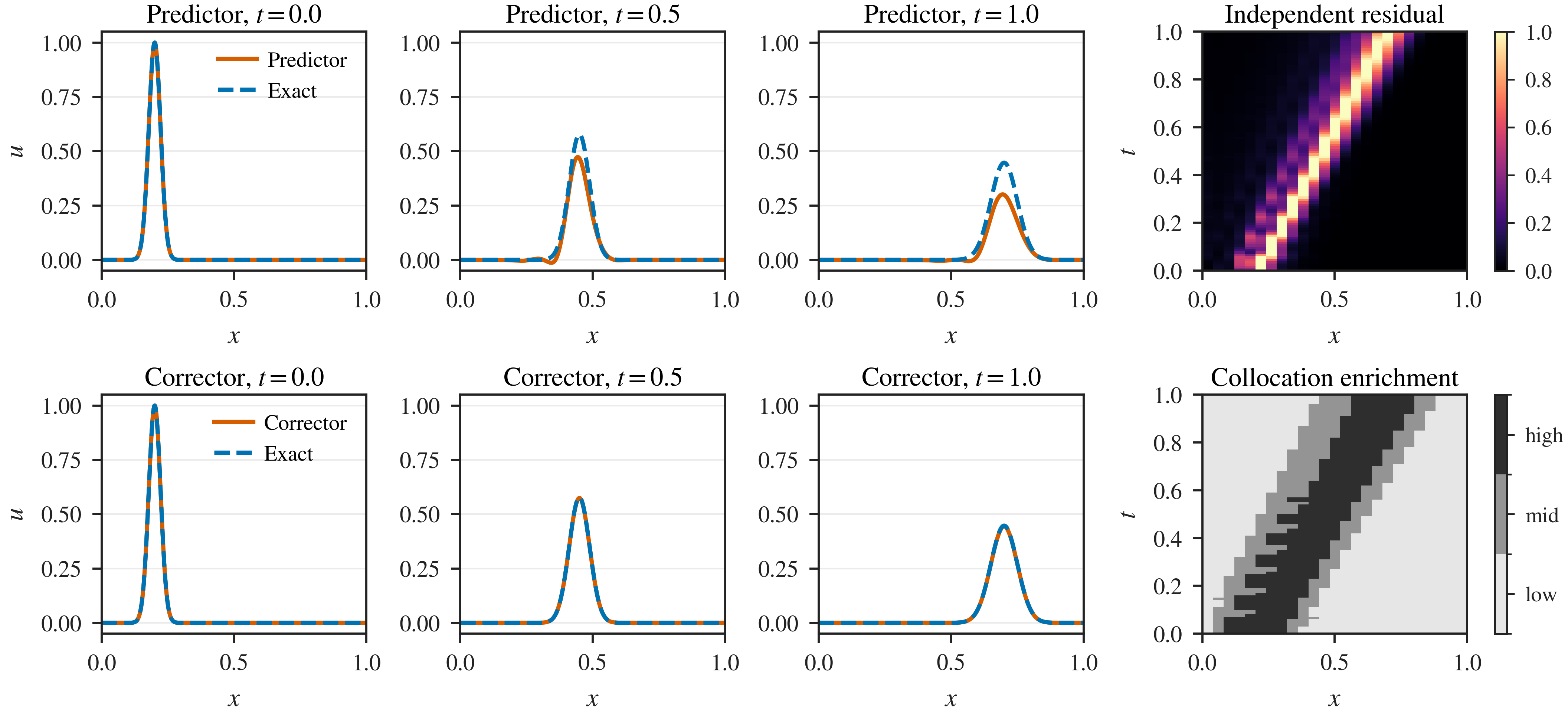}
	\caption{CFKS results for the advection--diffusion problem.
		Top row: predictor solutions at representative times and
		the independently evaluated PDE residual. Bottom row:
		adaptive corrector solutions and the corresponding space--time
		collocation enrichment. The enriched region follows the
		translating and diffusing Gaussian pulse.}
	\label{fig:advdiff_results}
\end{figure*}

We first consider two time-dependent transport problems to examine
the performance of sequential space--time-slab marching and
residual-adaptive collocation. The governing equations, computational
domains, and reference solutions are summarized in
Table~\ref{tab:test_cases}. Both experiments employ a deterministic
coarse predictor followed by a residual-guided corrector, with the
continuum-feature mean kernel retained between the two stages.
Independent residual evaluations are used to identify regions of
under-resolution and redistribute the PDE collocation points
accordingly. Figures~\ref{fig:advection_results} and
\ref{fig:advdiff_results} illustrate the resulting predictor and
corrector solutions, residual distributions, and space--time
collocation patterns.

For the linear advection problem, a narrow Gaussian pulse propagates
periodically across the spatial domain $[-1,1]$ with unit advection
speed. The coarse predictor captures the overall motion of the pulse
but accumulates visible phase and amplitude errors during propagation.
At the final time $t=1$, the predictor has a root-mean-square error
(RMSE) of $6.53\times10^{-2}$ and a maximum absolute error of
$3.73\times10^{-1}$. The independently evaluated residual identifies
a narrow moving region of under-resolution, producing a high-attention
space--time band aligned with the pulse trajectory, as shown in
Fig.~\ref{fig:advection_results}. Residual-guided collocation
enrichment follows this band and reduces the final RMSE to
$2.87\times10^{-3}$ and the maximum absolute error to
$2.13\times10^{-2}$. The corrector uses an average of 307.0 PDE
collocation points per slab, with a minimum of 288 and a maximum
of 320 points. The improved agreement with the exact solution
demonstrates that localized enrichment can resolve the transported
feature without changing the prescribed continuum-feature kernel.

The computational timings distinguish the cost of the kernel
solution from the additional operations required for residual
diagnostics and sequential transfer. For the advection predictor,
kernel assembly, spectral factorization, and coefficient solution
over all slabs require approximately $0.085$, $0.119$, and
$0.054$~s, respectively, giving a core CFKS time of $0.259$~s.
Residual probing and slab prediction/transfer require approximately
$19.48$ and $12.70$~s, respectively, and therefore dominate the
predictor-stage computational cost. For the adaptive corrector,
the core CFKS time increases to $4.64$~s, including $0.103$~s
for full operator-kernel assembly and $4.30$~s for cached spectral
factorizations associated with 52 distinct refinement patterns.
The total Stage-2 elapsed time, including prediction and transfer,
is $20.16$~s. These measurements indicate that the cost of
independent residual evaluation and dense solution transfer can
substantially exceed that of the underlying population-kernel
coefficient solve.

The advection--diffusion problem considers a translating Gaussian
pulse on $[0,1]$, with advection coefficient $a=0.5$ and diffusion
coefficient $\nu=10^{-3}$. In contrast to pure advection, the exact
solution broadens and decreases in amplitude as diffusion acts on
the moving pulse. The coarse predictor reproduces the overall
translation and spreading but increasingly underestimates the
pulse amplitude, yielding a final RMSE of $4.36\times10^{-2}$
and a maximum absolute error of $1.51\times10^{-1}$. As shown in
Fig.~\ref{fig:advdiff_results}, the independent residual again
identifies the moving region of strongest local variation, while
the enriched collocation band broadens consistently with the
diffusive spreading of the solution. The adaptive corrector
reduces the final RMSE to $9.42\times10^{-4}$ and the maximum
absolute error to $3.46\times10^{-3}$. The corrector employs an
average of 326.2 PDE points per slab, with the number varying
between 304 and 344 according to the residual-guided refinement
pattern.

For the advection--diffusion predictor, kernel assembly, spectral
factorization, and coefficient solution require approximately
$0.061$, $0.047$, and $0.048$~s, respectively, resulting in a
core CFKS time of $0.156$~s. Residual probing requires
$31.98$~s, while slab prediction and transfer require
$20.13$~s. The adaptive corrector has a core CFKS time of
$3.10$~s, of which $2.69$~s is spent on cached spectral
factorizations for 52 distinct refinement patterns. The total
Stage-2 elapsed time is $30.80$~s. Together with the pure
advection results, these measurements show that the numerical
cost of the slab-based computations is influenced strongly by
residual diagnostics, prediction-grid evaluation, and solution
transfer, rather than solely by assembly and solution of the
operator-kernel systems.

Taken together, the two transport experiments demonstrate that
residual-adaptive CFKS can concentrate collocation resolution along
moving localized structures while preserving the underlying
continuum-feature representation. In the pure advection case, the
enrichment follows the narrow trajectory of the periodically
transported pulse, whereas in the advection--diffusion case it
adapts to a translating and broadening profile. The accuracy
improvements are obtained by changing the locations and number
of discrete PDE constraints rather than introducing additional
finite Gaussian-RBF features or sampling new feature centers.
This distinction is important because the adaptive procedure
operates on the discretization of the governing equations while
the analytically prescribed mean kernel remains unchanged.
The results also indicate that improving the efficiency of
residual probing and slab-to-slab transfer is a relevant direction
for reducing the overall computational cost of time-dependent
CFKS implementations.

Closely related sharp-gradient transport benchmarks were previously
used to motivate the distributed physics-informed extreme learning
machine (PIELM) formulation \citep{DWIVEDI2021299}, in which spatial
domain decomposition was introduced to improve the resolution of
localized solution structures. In contrast, the present CFKS
computations retain a spatially global continuum-feature kernel
representation within each time slab, without introducing artificial
spatial interface conditions. Localized structures are instead
resolved by redistributing PDE collocation points according to
independently evaluated residual information. The resulting
calculations illustrate how deterministic population-kernel
representations can be combined with adaptive collocation without
requiring spatial domain decomposition.

\subsection{Viscous Burgers equation}
\label{sec:burgers_results}

We next consider the nonlinear viscous Burgers equation to examine
the applicability of CFKS beyond linear differential operators.
The problem is solved directly at viscosity $\nu=0.01$ over
$[-1,1]\times[0,1]$, with initial condition
$u(x,0)=-\sin(\pi x)$ and homogeneous Dirichlet conditions
at both spatial boundaries. The numerical solution is compared
with a Cole--Hopf reference solution. At this viscosity,
nonlinear steepening produces a sharp internal transition near
$x=0$, providing a demanding test of the continuum-feature
representation, analytical kernel derivatives, and nonlinear
residual minimization. The calculation employs a two-stage
predictor--corrector strategy combining prescribed multiscale
kernel refinement with residual-guided collocation enrichment.
Unlike the linear transport experiments, both the collocation
distribution and the Gaussian kernel widths are modified
between the two stages.

The temporal interval is divided into 40 uniform space--time
slabs, each of duration $\Delta t=0.025$, while the spatial
interval is partitioned into 20 equal blocks for residual
assessment and collocation refinement. On each slab, the
analytical trial-function embedding in
\eqref{eq:burgers_embedding} preserves the incoming solution
profile and homogeneous boundary conditions. After completing
a slab, the terminal solution is evaluated on a uniform
spatial transfer grid containing 1201 points, with the
prescribed boundary values imposed at the endpoints.
A cubic-spline interpolant of this terminal profile supplies
the incoming solution and its spatial derivatives for the
following slab. The predictor and corrector are marched
separately through the complete temporal interval, each
starting from the prescribed initial condition. The
predictor residual information determines the corrector
collocation distribution, but the predictor solution is
not imposed as the incoming profile of the corrector.

The nonlinear approximation uses the normalized global--local
continuum-feature mean kernel defined in
\eqref{eq:multiscale_mean_kernel}, with prescribed mixing
weight $\eta=0.30$. The global Gaussian center distribution
covers the spatial interval $[-1.3,1.3]$, whereas the local
component uses the narrower interval $[-0.3,0.3]$ to provide
additional resolution around the developing internal transition.
The corresponding local-time center intervals are
$[-0.04,\Delta t+0.04]$ for the global component and
$[-0.035,\Delta t+0.035]$ for the local component. Each
kernel component is normalized by its diagonal value at
the spatial origin and the midpoint of the local time
slab. For the Stage-1 predictor, the prescribed spatial
and temporal Gaussian widths are $(0.20,0.04)$ for the
global kernel and $(0.06,0.06)$ for the local kernel.
In Stage~2, these widths are reduced to $(0.10,0.02)$
and $(0.015,0.015)$, respectively. The prescribed kernel
parameters remain fixed across the slabs within each
stage but differ between the predictor and corrector
computations.

The nonlinear coefficient solution on each slab begins
with a frozen-advection linearization based on the
incoming solution profile, followed by a Picard update.
The linearized coefficient systems are solved using
singular-value-decomposition (SVD) regularization, with
relative ridge factors of $10^{-8}$ and $10^{-9}$ for
the frozen-advection and Picard steps, respectively.
The resulting coefficients initialize nonlinear
least-squares minimization of the full Burgers residual,
using the analytical Jacobian derived in
\eqref{eq:burgers_jacobian}. The nonlinear solver employs
a trust-region reflective algorithm with an iterative
least-squares linear solver, a maximum of 20 function
evaluations, and convergence tolerances of $10^{-8}$
for the objective function, coefficient changes,
and gradient norm. This initialization and optimization
procedure is used in both stages, so the distinction
between the predictor and corrector lies principally
in their prescribed kernel scales and collocation
distributions.

For the Stage-1 predictor, each of the 20 spatial
blocks contains 10 PDE collocation points, giving
200 points per time slab. After completing the predictor
march, the nonlinear PDE residual is evaluated on
independent probe grids within each block. Each probe
grid contains 11 spatial locations and 7 local-time
locations, giving 77 residual evaluation points per
block. The spatial probe locations extend from
5\% to 95\% of the block width, while the temporal
locations range from $0.08\Delta t$ to
$0.96\Delta t$. Residual root-mean-square and maximum
magnitudes are evaluated separately for each block,
and the normalized indicator in
\eqref{eq:residual_indicator} is computed using
$\omega=0.7$ and $\varepsilon=10^{-14}$.
The normalization is performed independently within
each time slab, allowing the refinement decisions
to respond to the spatial distribution of residuals
at different stages of the evolution.

The refinement thresholds are prescribed as
$s_{\mathrm{mid}}=0.10$ and
$s_{\mathrm{high}}=0.35$, defining low-, medium-,
and high-attention regions. To avoid abrupt changes
in collocation density, medium- and high-attention
assignments are extended to their immediate
neighboring spatial blocks without wrapping around
the physical boundaries. High-attention assignments
take precedence wherever the expanded regions overlap.
The corresponding collocation budgets are
$N_{\mathrm{low}}=10$,
$N_{\mathrm{mid}}=20$, and
$N_{\mathrm{high}}=30$ PDE points per block, with
deterministic collocation locations generated using
a two-dimensional Halton sequence with bases 2 and 3.
Within each block, the selected number of Halton
points is mapped to the corresponding spatial and
local-time interval. Since the kernel representer
sites coincide with the PDE collocation locations,
residual-guided enrichment changes both the number
of PDE constraints and the dimension of the nonlinear
coefficient system. The resulting Stage-2 corrector
uses between 350 and 600 PDE points per slab, with
an average of 402.75 points.

\begin{figure*}[t]
	\centering
	\includegraphics[width=\textwidth]
	{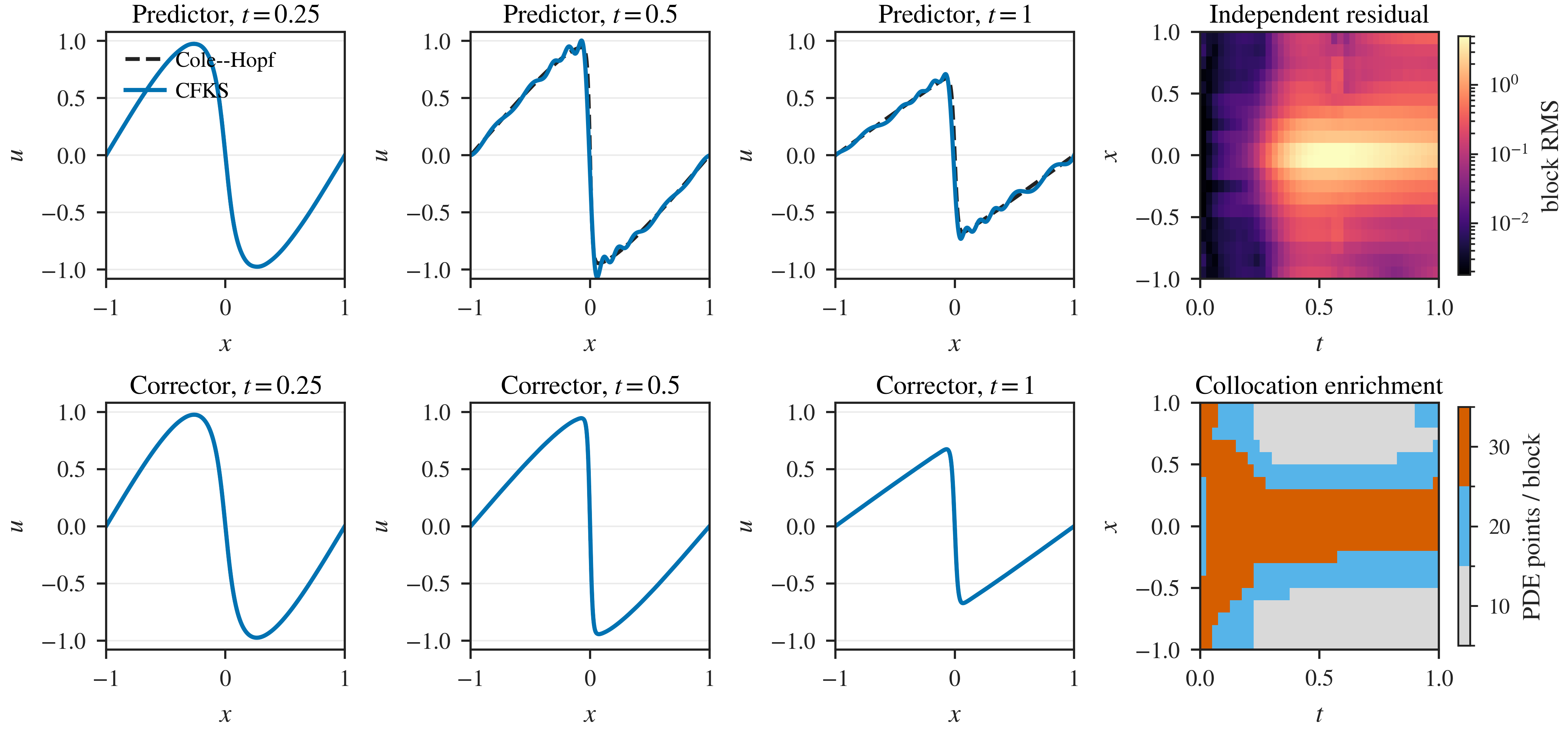}
	\caption{Two-stage multiscale CFKS solution of viscous
		Burgers' equation at $\nu=0.01$. Top row: coarse
		predictor solutions at $t=0.25$, $0.5$, and $1$,
		together with the independently evaluated nonlinear
		PDE residual. Bottom row: adaptive corrector solutions
		at the corresponding times and the residual-guided
		collocation distribution. Dashed curves denote the
		Cole--Hopf reference solution. The corrector combines
		increased collocation resolution near the developing
		internal transition with narrower prescribed
		multiscale kernel widths.}
	\label{fig:burgers_residual_maps}
\end{figure*}

Figure~\ref{fig:burgers_residual_maps} illustrates
the two-stage solution process and its associated
collocation refinement. The coarse predictor initially
follows the Cole--Hopf reference solution but develops
oscillatory errors as the nonlinear internal transition
becomes sharper. These discrepancies are especially
pronounced near the central gradient at later times,
where the independently evaluated PDE residual also
identifies increased under-resolution. The Stage-2
corrector allocates additional collocation points
around the developing transition while using the
narrower prescribed multiscale kernel widths. The
corrected solution exhibits substantially improved
agreement with the reference, indicating that the
combined refinement strategy can resolve localized
nonlinear structures more accurately than the
coarse predictor.

The quantitative errors at four representative times
are summarized in Table~\ref{tab:burgers_errors}.
At $t=0.25$, the predictor RMSE is
$9.02\times10^{-4}$, which decreases to
$2.21\times10^{-4}$ after correction. As the
internal transition develops, the difference
between the two stages becomes considerably
more pronounced. At $t=0.5$, the predictor
RMSE of $4.65\times10^{-2}$ is reduced to
$3.44\times10^{-4}$, while at $t=0.75$
the corresponding values are
$5.97\times10^{-2}$ and
$4.70\times10^{-4}$. At the final time
$t=1$, the predictor has an RMSE of
$4.83\times10^{-2}$ and a maximum
absolute error of $3.01\times10^{-1}$,
whereas the corrector reduces these
values to $4.55\times10^{-4}$ and
$1.81\times10^{-3}$, respectively.
The final-time RMSE is therefore reduced
by approximately a factor of 106.

\begin{table}[t]
	\centering
	\caption{Predictor and corrector errors for viscous
		Burgers' equation at $\nu=0.01$, evaluated against
		the Cole--Hopf reference solution.}
	\label{tab:burgers_errors}
	\footnotesize
	\setlength{\tabcolsep}{4pt}
	\renewcommand{\arraystretch}{1.15}
	\begin{tabular}{ccccc}
		\hline
		& \multicolumn{2}{c}{Predictor}
		& \multicolumn{2}{c}{Corrector} \\
		\cline{2-3}\cline{4-5}
		$t$
		& RMSE & $L^\infty$
		& RMSE & $L^\infty$ \\
		\hline
		$0.25$
		& $9.02\times10^{-4}$
		& $4.69\times10^{-3}$
		& $2.21\times10^{-4}$
		& $9.64\times10^{-4}$ \\
		$0.50$
		& $4.65\times10^{-2}$
		& $3.32\times10^{-1}$
		& $3.44\times10^{-4}$
		& $2.32\times10^{-3}$ \\
		$0.75$
		& $5.97\times10^{-2}$
		& $4.16\times10^{-1}$
		& $4.70\times10^{-4}$
		& $2.65\times10^{-3}$ \\
		$1.00$
		& $4.83\times10^{-2}$
		& $3.01\times10^{-1}$
		& $4.55\times10^{-4}$
		& $1.81\times10^{-3}$ \\
		\hline
	\end{tabular}
\end{table}

\begin{figure*}[t]
	\centering
	\includegraphics[width=0.78\textwidth]
	{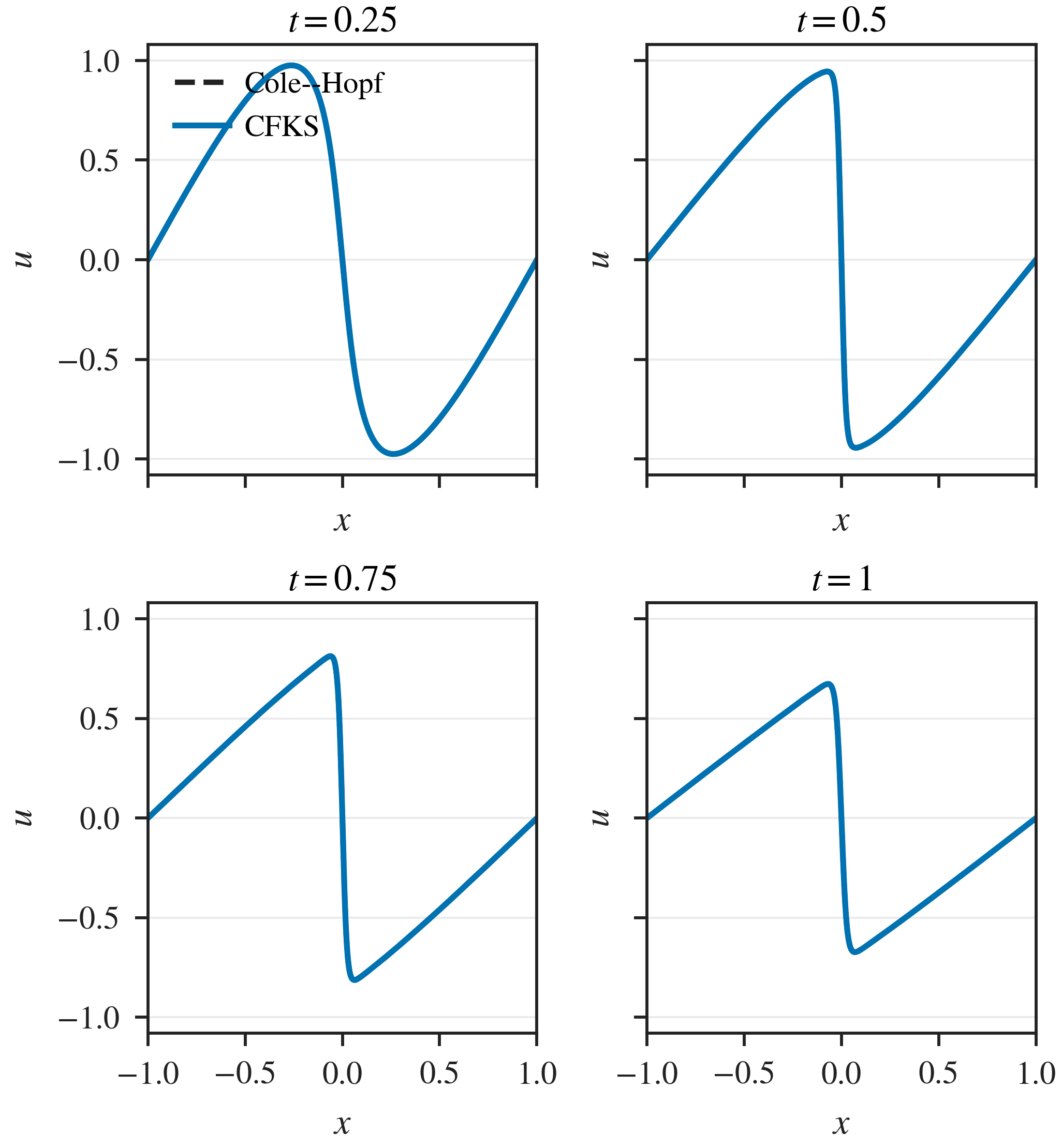}
	\caption{Corrected multiscale CFKS solution of
		viscous Burgers' equation at $\nu=0.01$
		compared with the Cole--Hopf reference
		solution at $t=0.25$, $0.5$, $0.75$,
		and $1$. The corrected solution closely
		follows the reference while resolving
		the developing internal transition
		near $x=0$.}
	\label{fig:burgers_solutions}
\end{figure*}

Figure~\ref{fig:burgers_solutions} provides a
further comparison between the corrected CFKS
solution and the Cole--Hopf reference at
$t=0.25$, $0.5$, $0.75$, and $1$. The
corrected approximation remains in close
agreement with the reference throughout the
simulation, including the increasingly steep
internal transition near $x=0$. Both the
amplitude and position of this localized
structure are reproduced without spatial
domain decomposition or viscosity continuation.
The improvement should, however, be interpreted
as the combined effect of residual-guided
collocation enrichment and refinement of
the prescribed Gaussian kernel widths,
since these two changes are introduced
simultaneously in the corrector.

Overall, the Burgers experiment demonstrates that the
continuum-feature mean-kernel representation can be
extended to nonlinear PDEs through linearized
initialization and nonlinear least-squares residual
minimization. Analytical kernel derivatives provide
the nonlinear PDE residual and its Jacobian, while
the global--local multiscale representation supplies
additional flexibility for resolving localized
solution structures. The two-stage procedure
improves accuracy by combining residual-guided
collocation enrichment with prescribed kernel-scale
refinement. The computation is performed directly
at the target viscosity without introducing
finite Gaussian-RBF center sampling, viscosity
continuation, or spatial domain decomposition.

\subsection{Finite-feature to continuum convergence}
\label{sec:finite_to_continuum}

To verify the finite-feature interpretation of CFKS, we use the first-order
oscillatory ODE benchmark defined in Sec.~\ref{sec:test_cases},
\begin{equation}
	u'(x)=\cos(15x),
	\qquad
	x\in[-2\pi,2\pi],
	\qquad
	u(0)=0,
\end{equation}
with exact solution
$u_{\mathrm{ex}}(x)=\frac{1}{15}\sin(15x)$. This case provides a convenient
controlled setting in which the finite Gaussian-RBF feature representation
and the corresponding CFKS population limit can be compared directly.

\begin{figure*}[t]
	\centering
	
	\begin{minipage}[t]{0.49\textwidth}
		\centering
		\includegraphics[width=\linewidth]
		{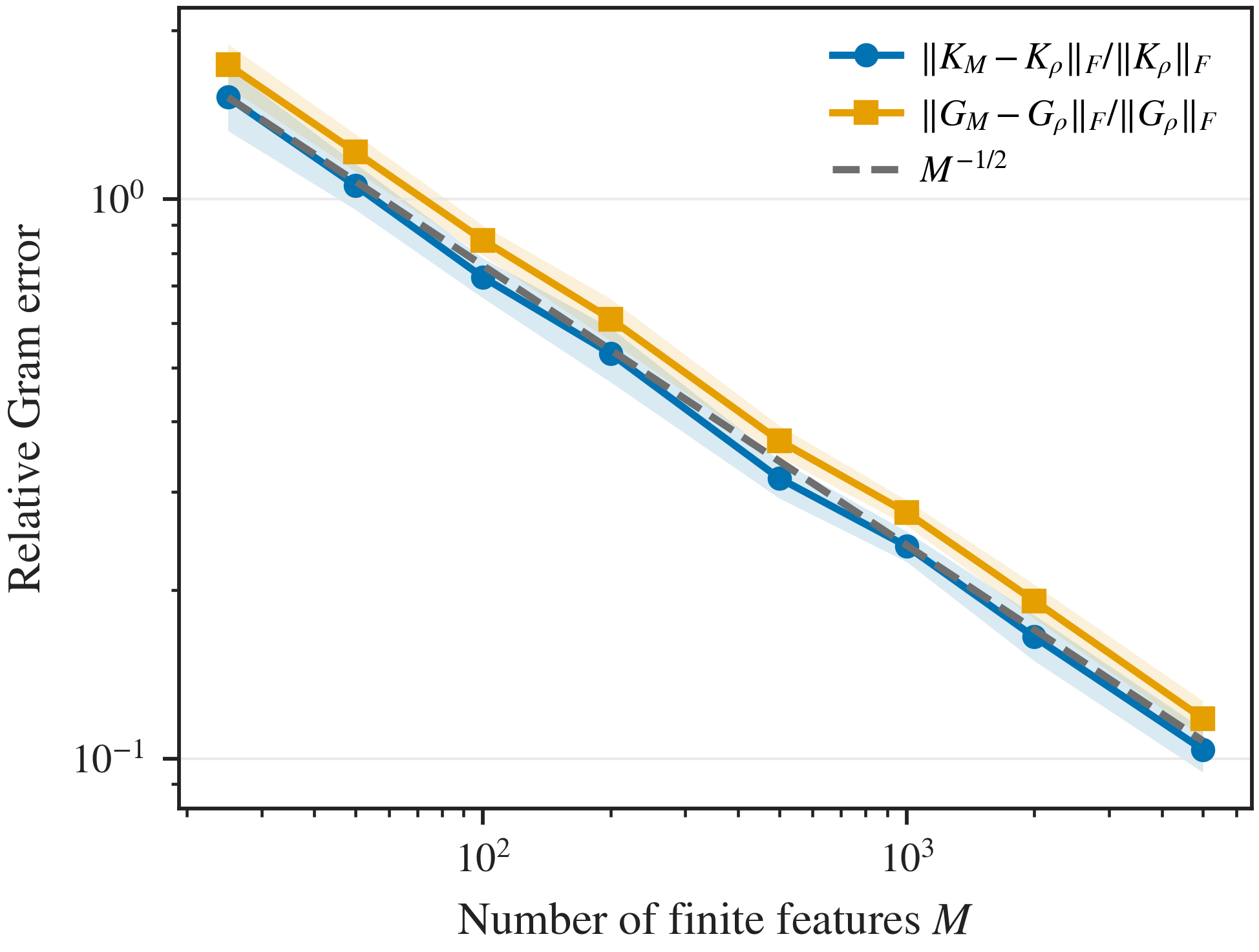}
	\end{minipage}
	\hfill
	\begin{minipage}[t]{0.49\textwidth}
		\centering
		\includegraphics[width=\linewidth]
		{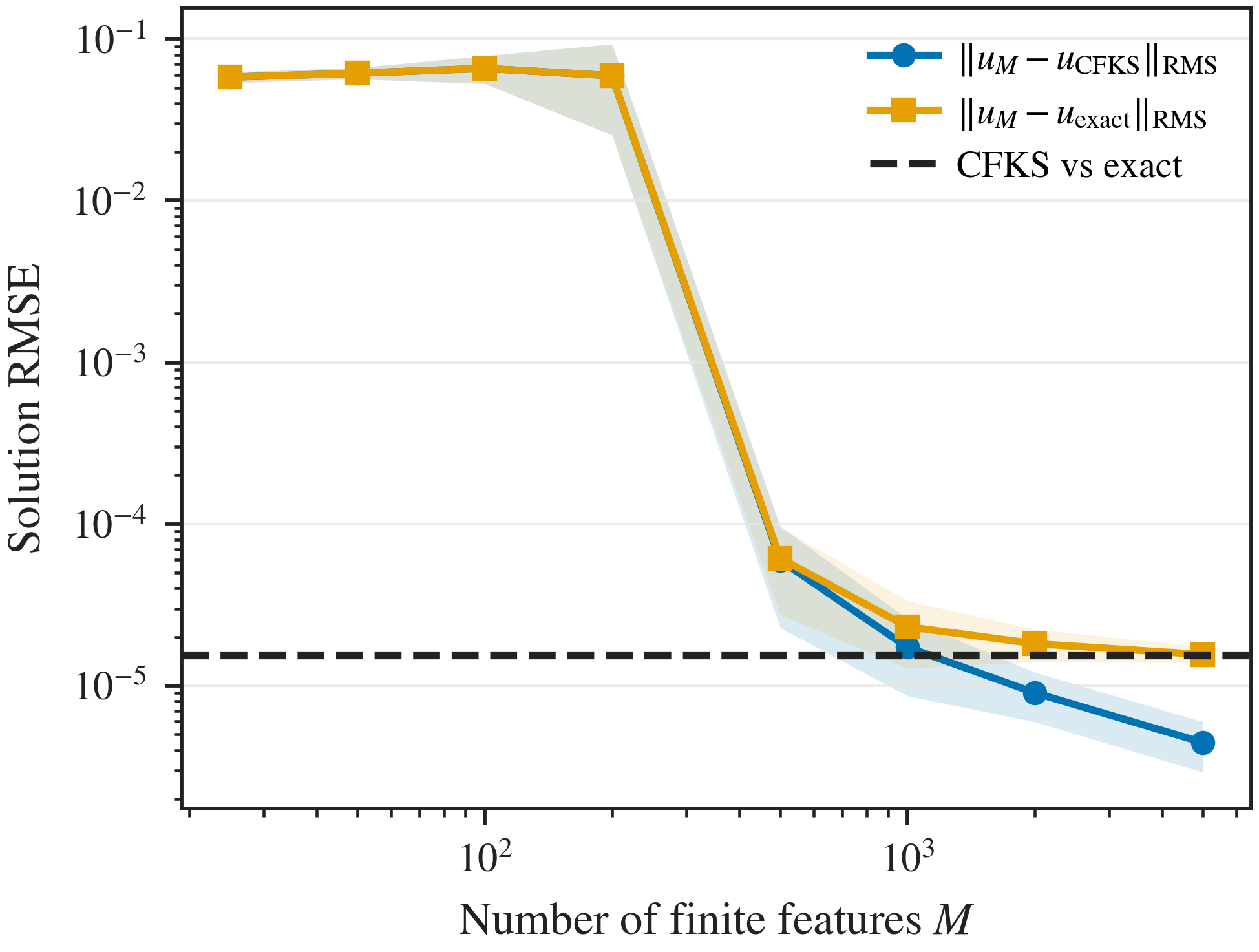}
	\end{minipage}
	
	\caption{Finite-feature convergence toward CFKS. Left: relative errors
		of the empirical value and operator Gram matrices as the number of
		sampled Gaussian RBF features $M$ increases; shaded bands denote
		realization-to-realization variability and the dashed line indicates
		the $M^{-1/2}$ reference rate. Right: RMS difference between the
		finite-feature and CFKS solutions, together with the finite-feature
		error relative to the exact solution. The dashed horizontal line denotes
		the CFKS error relative to the exact solution.}
	\label{fig:finite_feature_convergence}
\end{figure*}

Figure~\ref{fig:finite_feature_convergence} verifies numerically that the
finite Gaussian-RBF feature representation approaches the CFKS population
limit as the number of sampled features $M$ increases. The top panel shows
the relative Frobenius errors of the empirical value Gram matrix,
$\|K_M-K_\rho\|_F/\|K_\rho\|_F$, and of the operator-applied Gram matrix,
$\|G_M-G_\rho\|_F/\|G_\rho\|_F$. Both decrease systematically with $M$ and
closely follow the reference $M^{-1/2}$ slope expected for Monte Carlo
sampling of the center population. The mean value-kernel error decreases
from $1.52$ at $M=25$ to $1.04\times10^{-1}$ at $M=5000$, while the
operator-Gram error decreases from $1.74$ to $1.18\times10^{-1}$ over the
same range. The shrinking realization-to-realization spread further confirms
convergence toward a deterministic population kernel.

The bottom panel shows the corresponding effect at the solution level. For
small feature counts, the finite-feature approximation remains far from the
CFKS solution, with RMS differences of approximately $6\times10^{-2}$ for
$M\leq200$. A marked transition occurs once the feature space becomes
sufficiently rich: at $M=500$ the mean finite-feature-to-CFKS RMSE falls to
$5.98\times10^{-5}$, and subsequently decreases to $4.45\times10^{-6}$ at
$M=5000$. The CFKS solution itself has an RMSE of
$1.54\times10^{-5}$ relative to the exact solution. At large $M$, the
finite-feature solutions therefore converge toward the deterministic CFKS
reference, while their realization-dependent variability also decreases.

These results provide direct numerical evidence for the interpretation of
CFKS as the population limit of the underlying finite Gaussian-RBF feature
model. The continuum formulation removes the finite-feature sampling error
and the associated realization variability, while retaining the operator
Gram structure obtained from the original RBF representation.
\subsection{Conditioning and regularization}
\label{sec:conditioning}

\begin{figure*}[t]
	\centering
	
	\begin{minipage}[t]{0.49\textwidth}
		\centering
		\includegraphics[width=\linewidth]
		{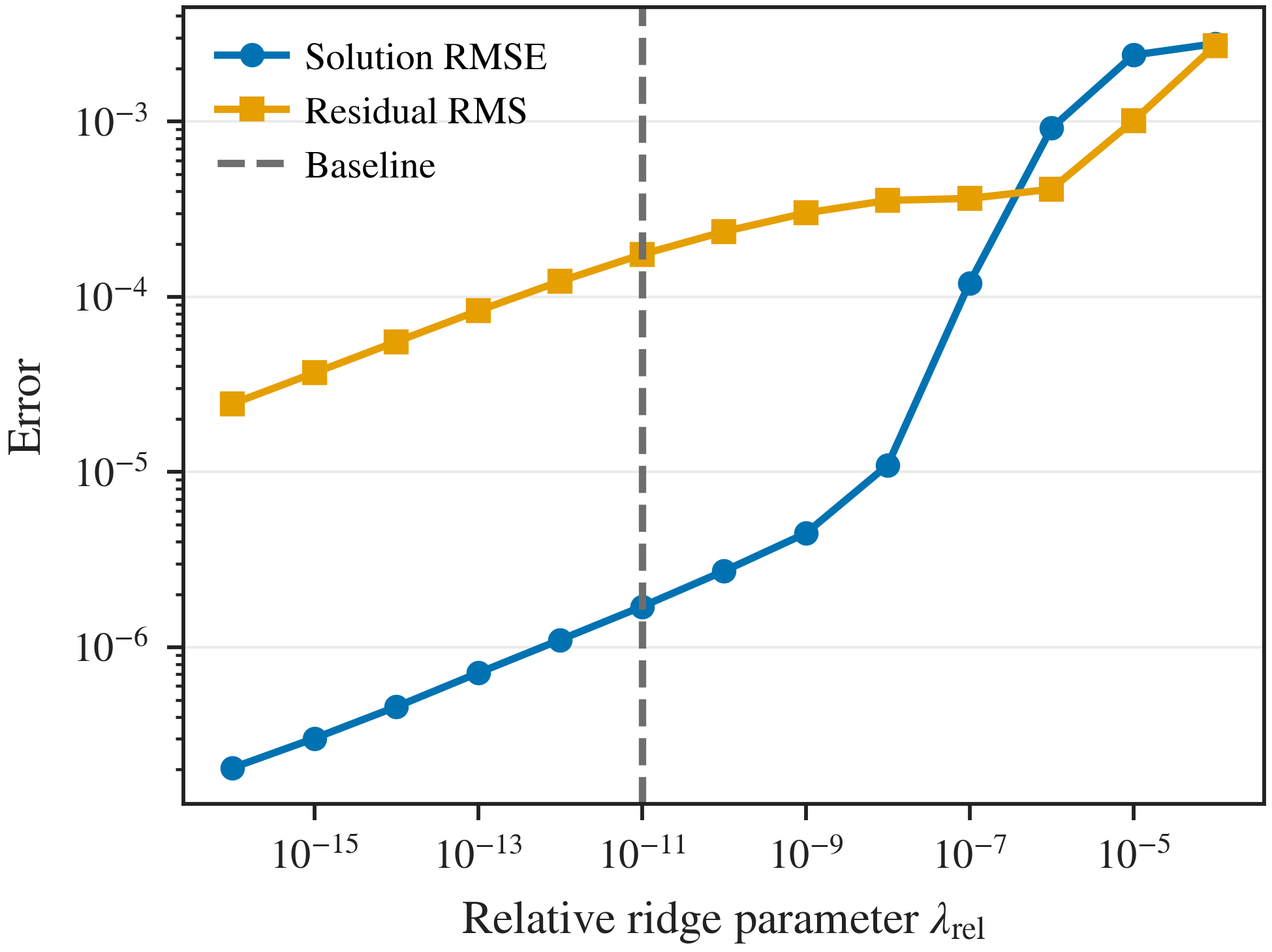}
	\end{minipage}
	\hfill
	\begin{minipage}[t]{0.49\textwidth}
		\centering
		\includegraphics[width=\linewidth]
		{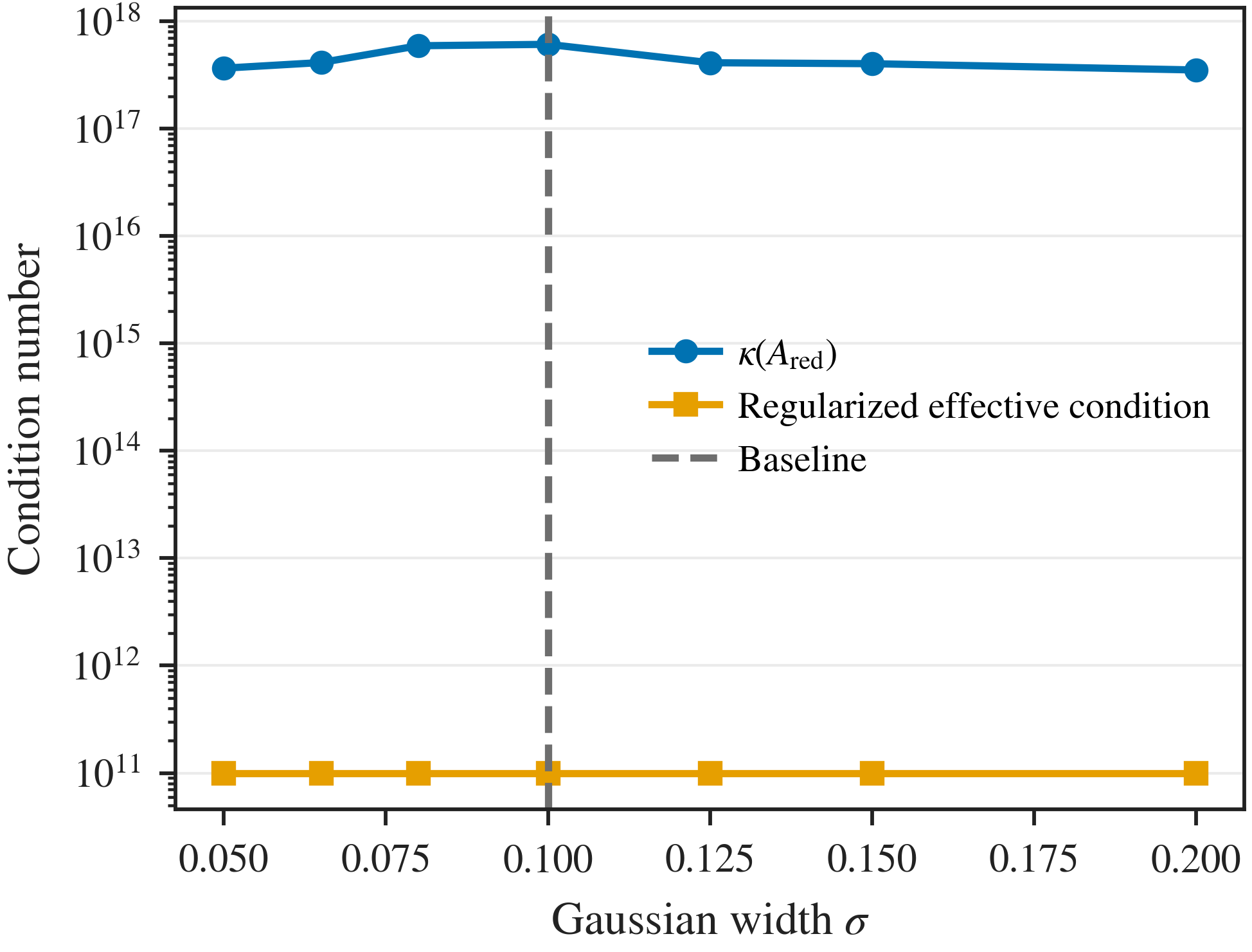}
	\end{minipage}
	
	\caption{Conditioning and regularization study for the first-order
		oscillatory ODE benchmark. Left: dependence of the CFKS solution RMSE
		and PDE-residual RMS on the relative Tikhonov parameter
		$\lambda_{\rm rel}$ at fixed $\sigma=0.10$. Right: raw reduced-system
		condition number and regularized effective condition number as the
		Gaussian width $\sigma$ is varied. Dashed vertical lines indicate the
		baseline parameters used in the numerical experiments.}
	\label{fig:conditioning_regularization}
\end{figure*}

Figure~\ref{fig:conditioning_regularization} examines the sensitivity of
CFKS to the two principal numerical parameters using the first-order
oscillatory ODE benchmark. The regularization sweep shows the expected
trade-off between stabilization and approximation accuracy: insufficient
regularization leaves the solve sensitive to small singular values, whereas
excessive regularization increases both the solution and residual errors.
The baseline value $\lambda_{\rm rel}=10^{-11}$ lies within the stable
low-error region.

The lower panel shows that the Gaussian width $\sigma$ has a strong effect
on the conditioning of the reduced operator system. Increasing $\sigma$
changes the overlap among neighboring population-kernel basis functions and
can substantially increase the raw condition number. Tikhonov
regularization reduces this effective conditioning burden while preserving
the solution accuracy over a useful range of kernel widths. These results
motivate the moderate width and weak regularization used throughout the
remaining experiments.

\subsection{Advection with two separated Gaussian pulses}
\label{sec:advection_two_gaussians}

To further test the spatial selectivity of the residual-adaptive procedure,
we consider the same periodic advection equation as in
Sec.~\ref{sec:test_cases}, but replace the single Gaussian initial condition
with two separated pulses,
\begin{equation}
	u(x,0)
	=
	\exp\!\left[-\left(25(x+0.4)\right)^2\right]
	+
	\exp\!\left[-\left(25(x-0.4)\right)^2\right].
\end{equation}
The two pulses therefore have identical width and are initially centered at
$x=-0.4$ and $x=0.4$. The exact periodic solution is obtained by translating
the complete initial profile with unit advection speed.

\begin{figure*}[t]
	\centering
	\includegraphics[width=\textwidth]{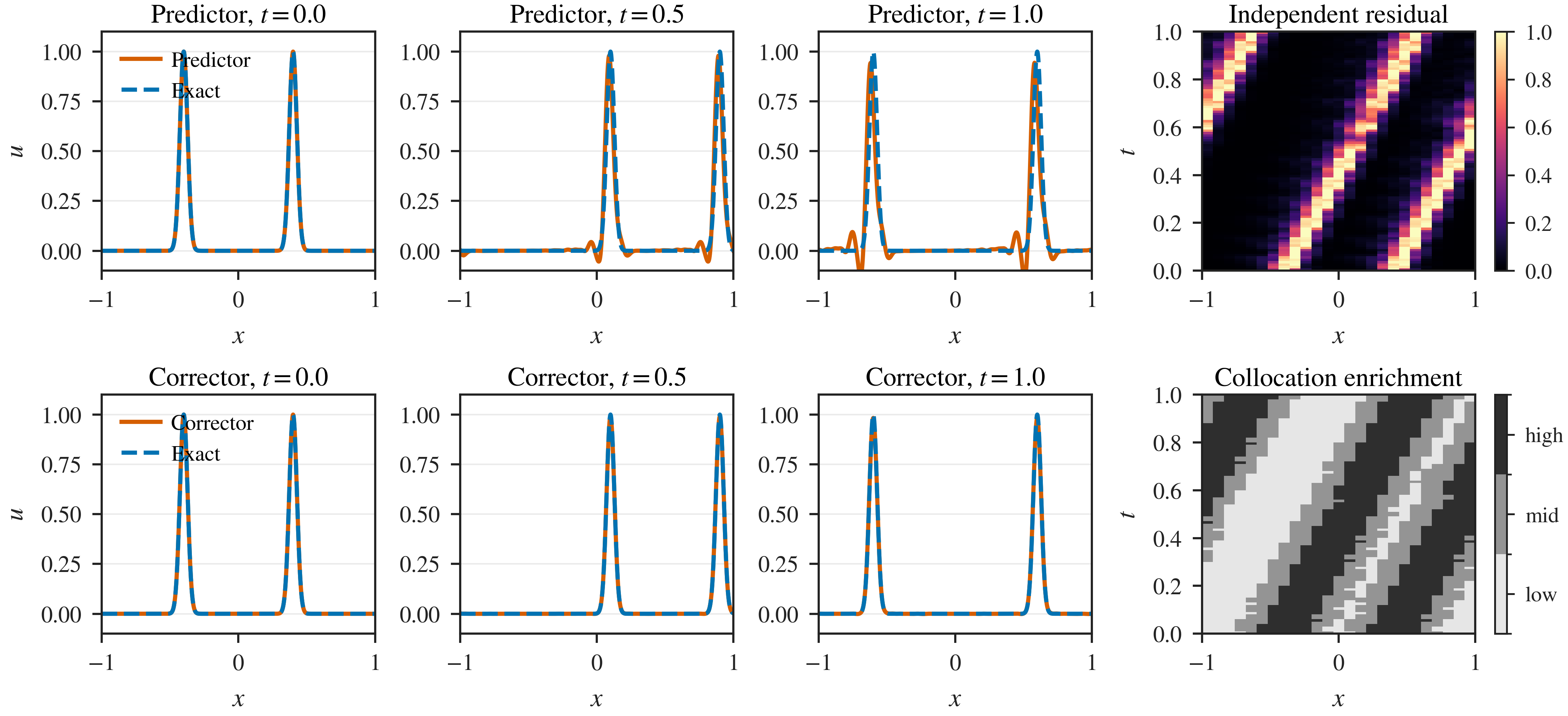}
	\caption{Periodic advection of two separated Gaussian pulses.
		Top row: predictor solution at $t=0$, $0.5$, and $1.0$, followed by the
		independently evaluated residual over the space--time domain.
		Bottom row: adaptive-corrector solution at the same times and the
		resulting collocation-enrichment pattern. The residual traces both
		propagating pulses and directs increased collocation density along their
		corresponding space--time trajectories.}
	\label{fig:advection_two_gaussians}
\end{figure*}

Figure~\ref{fig:advection_two_gaussians} shows that the predictor captures
the transport of both pulses but develops localized oscillatory errors near
the moving sharp features. The independent residual identifies two distinct
space--time trajectories associated with the advecting pulses, and the
corrector consequently concentrates additional collocation points along both
paths. The final-time predictor error is
$\mathrm{RMSE}=9.16\times10^{-2}$ with a maximum error of
$3.74\times10^{-1}$. After residual-guided enrichment, these errors decrease
to $\mathrm{RMSE}=3.31\times10^{-3}$ and a maximum error of
$1.84\times10^{-2}$, respectively.

The adaptive corrector uses an average of $413.0$ PDE collocation points per
slab, with a range of $384$--$432$. Thus, the residual-based enrichment is
not tied to a single localized feature; it can identify and track multiple
simultaneously propagating structures while leaving the underlying CFKS
kernel and physical constraints unchanged.

\subsection{Overall observations}
\label{sec:overall_observations}

The numerical experiments demonstrate the applicability of CFKS across
differential problems involving oscillatory solutions, multiscale
structures, irregular geometries, higher-order operators, localized
boundary layers, time-dependent transport, and nonlinear dynamics.
The finite-feature convergence study provides direct numerical
evidence that empirical Gaussian-RBF value and operator Gram matrices
approach their analytically defined continuum counterparts as the
number of sampled features increases. This establishes the connection
between finite Gaussian-RBF representations and the deterministic
continuum-feature formulation used throughout the paper. The ODE,
Poisson, and biharmonic benchmarks further demonstrate that the
resulting mean kernel and its analytical derivatives can be employed
for differential operators of different orders without introducing
separate finite feature realizations. The successful treatment of
the irregular five-petal domain also illustrates that the prescribed
kernel center-support region need not coincide with the physical
computational domain, since the domain geometry and physical
conditions enter through the corresponding constraint functionals.

The singularly perturbed and time-dependent transport experiments
demonstrate the role of residual-adaptive collocation as an additional
resolution strategy within the continuum-feature framework.
Independent PDE residual evaluations identify localized regions
where the predictor approximation is insufficiently resolved,
and the corrector redistributes collocation points accordingly.
The singularly perturbed problem shows that concentrating additional
constraints near a narrow boundary layer can be more effective than
uniform refinement with the same point budget. Similarly, the
advection and advection--diffusion experiments demonstrate that
residual-guided enrichment can follow moving localized structures,
while the two-pulse advection benchmark confirms that multiple
separated regions of under-resolution can be identified
simultaneously. In these linear examples, the prescribed
continuum-feature mean kernel remains unchanged between the
predictor and corrector stages; refinement modifies the
discrete collocation representation rather than the underlying
Gaussian feature distribution.

The viscous Burgers experiment extends the numerical assessment
to a nonlinear differential operator. In this setting, the
continuum-feature mean kernel defines a deterministic trial
representation, while the unknown coefficients are determined
through nonlinear least-squares residual minimization.
Linearized initialization and analytical kernel derivatives
provide the starting approximation, nonlinear residual, and
corresponding Jacobian. At the investigated viscosity
$\nu=0.01$, the two-stage procedure substantially improves
the resolution of the developing internal transition.
Unlike the linear transport experiments, however, the
Burgers corrector combines residual-guided collocation
enrichment with narrower prescribed Gaussian widths in
a global--local multiscale kernel. The observed accuracy
improvement therefore reflects their combined effect,
and the present experiment does not separately quantify
the contribution of either refinement mechanism.

The conditioning and timing results also clarify the
computational characteristics of CFKS. The regularization
and Gaussian-width study for the first-order ODE confirms
that the accuracy and conditioning of the discrete
operator-kernel system depend on these prescribed
numerical parameters. In the steady linear benchmarks,
the main computational costs are associated with
analytical kernel assembly and dense linear algebra,
while null-space construction contributes comparatively
little to the measured solve times. For the slab-based
transport problems, residual evaluation, dense solution
prediction, and temporal transfer can require more
computational time than the core kernel coefficient
solves. Nonlinear problems introduce additional costs
through repeated residual and Jacobian evaluations
during iterative coefficient optimization. Consequently,
the overall computational efficiency depends on both
the kernel discretization and the numerical procedures
used for refinement, diagnostics, and time advancement.

Taken together, these findings support CFKS as a
deterministic computational framework connecting
finite Gaussian-RBF feature representations with
analytically constructed continuum-feature kernel
PDE solvers. Its principal representational advantage
is the elimination of the finite feature count and
individual center-sampling variability, rather than
the elimination of all numerical parameter choices.
Gaussian widths, bounded center distributions,
collocation configurations, regularization, and,
where applicable, nonlinear initialization remain
important aspects of the numerical formulation.
The experiments demonstrate the flexibility of
this approach across the considered benchmarks,
while also identifying parameter selection,
conditioning, and computational scalability
as important considerations for broader applications.

\paragraph{Remarks}

\begin{itemize}
	
	\item \textbf{Scope of nonlinear validation.}
	The nonlinear CFKS formulation is demonstrated
	numerically using viscous Burgers' equation at
	$\nu=0.01$. Although the analytical kernel
	derivatives and nonlinear least-squares
	formulation are applicable to other sufficiently
	smooth nonlinear differential operators, their
	numerical accuracy, robustness, and convergence
	behavior have not been systematically investigated
	across broader classes of nonlinear PDEs.
	The reported Burgers corrector modifies both
	the collocation distribution and the prescribed
	multiscale kernel widths, so their individual
	effects cannot be distinguished from the
	present results. Further numerical studies
	involving different nonlinear operators,
	parameter regimes, and separate refinement
	strategies are needed to assess the generality
	and limitations of the nonlinear extension.
	
	\item \textbf{Kernel parameters and numerical stability.}
	Although analytical marginalization removes
	finite-feature sampling variability, CFKS
	remains dependent on prescribed Gaussian
	widths, center distributions, and
	regularization parameters. The conditioning
	study illustrates the influence of Gaussian
	width and regularization on a representative
	linear problem, but does not establish a
	universally optimal parameter-selection rule.
	For problems containing localized structures,
	the appropriate kernel scales must also be
	considered in relation to the spatial and
	temporal resolution. In sequential
	space--time-slab computations, the accuracy
	of transferring the terminal solution
	between slabs provides an additional
	numerical consideration. Systematic
	parameter-selection procedures and
	conditioning-aware kernel refinement
	therefore remain worthwhile directions
	for further investigation.
	
	\item \textbf{Computational scalability.}
	CFKS avoids explicit finite Gaussian-RBF
	feature realizations but retains the
	computational demands of dense kernel
	discretizations. For a linear system
	involving $N$ constraints, storing the
	full operator Gram matrix requires
	$\mathcal{O}(N^2)$ memory, while direct
	dense factorization may require
	$\mathcal{O}(N^3)$ operations.
	Nonlinear formulations additionally
	require repeated evaluation of the
	PDE residual and its Jacobian.
	Sequential space--time-slab marching
	limits the size of individual temporal
	problems, and residual-adaptive
	collocation can concentrate resolution
	in selected regions, but neither
	mechanism removes the general
	scalability limitations of dense
	kernel systems. Low-rank kernel
	approximations, matrix-free iterative
	methods, efficient residual
	evaluation, and localized or
	domain-decomposed formulations
	are potential approaches for
	extending CFKS to larger-scale
	applications.
	
\end{itemize}

\section{Conclusions}
\label{sec:conclusion}

This work introduced Continuum-Feature Kernel Solver (CFKS), a
deterministic framework for solving differential equations through
analytical marginalization of a bounded continuum of Gaussian radial
basis function (RBF) features. By replacing a finite feature
realization with its exact continuum-feature mean kernel, CFKS
eliminates the need to prescribe a finite feature count or select
individual feature centers. The resulting kernel and its
differential-operator counterparts are evaluated analytically and
used directly in the numerical discretization. For linear problems,
this construction leads to symmetric operator-kernel collocation
systems, with boundary and initial conditions enforced through
constraint-compatible null-space parameterization. For nonlinear
problems, the continuum-feature representation is retained, while
the unknown coefficients are determined through linearized
initialization and nonlinear least-squares residual minimization
using analytical kernel derivatives and Jacobians.

The numerical investigations support the proposed formulation
from both the feature-representation and differential-equation
perspectives. Finite-feature convergence experiments demonstrate
that empirical Gaussian-RBF value and operator Gram matrices
approach their analytically derived continuum counterparts as
the number of sampled features increases. The corresponding
finite-feature solutions also approach the deterministic CFKS
solution, confirming the computational connection between
finite Gaussian-RBF models and the continuum-feature formulation.
The linear benchmarks demonstrate the applicability of the
analytical operator kernels to oscillatory and multiscale ODEs,
second- and fourth-order elliptic problems on irregular domains,
singularly perturbed boundary layers, and time-dependent
transport equations. Residual-guided collocation enrichment
further improves the resolution of localized solution structures
without changing the underlying mean kernel in the linear
adaptive examples. The transport experiments additionally
demonstrate that this strategy can track multiple moving
localized features within a spatially global kernel formulation.

The viscous Burgers experiment extends the numerical assessment
to a nonlinear differential operator. At the investigated
viscosity of $\nu=0.01$, the continuum-feature mean kernel,
analytical constraint embedding, and nonlinear residual
minimization provide an accurate approximation of the developing
internal transition. The two-stage predictor--corrector procedure
combines residual-guided collocation enrichment with narrower
prescribed global and local Gaussian kernel widths, producing
substantial improvements over the coarse predictor. These results
demonstrate the feasibility of extending CFKS to nonlinear PDEs
without introducing finite Gaussian-RBF feature realizations.
However, because both the collocation distribution and kernel
scales are modified between stages, the individual contributions
of these refinement mechanisms cannot be isolated from the
present experiment. Further assessment across different
nonlinear operators and parameter regimes is therefore needed.

The principal significance of CFKS lies in using the
continuum-feature limit as a direct computational representation
rather than solely as an asymptotic interpretation of a finite
feature model. Analytical marginalization removes finite-feature
sampling variability while preserving the connection to the
underlying Gaussian-RBF representation. Nevertheless, the method
retains prescribed Gaussian widths, bounded center distributions,
collocation configurations, and regularization parameters, whose
selection influences accuracy and conditioning. Moreover, CFKS
inherits the dense matrix assembly and linear-algebra costs of
global kernel methods, while nonlinear problems require additional
iterative computations. The time-dependent experiments also
highlight the computational expense associated with residual
evaluation and slab-to-slab solution transfer. Future research
will address scalable low-rank and iterative kernel formulations,
efficient residual evaluation, conditioning-aware parameter
selection, and localized or domain-decomposed extensions.
Broader nonlinear benchmarks and systematic investigations of
collocation and kernel-scale refinement will further clarify
the accuracy, robustness, and computational limits of CFKS.

\appendix
\section{From finite features to the CFKS representer form}
\label{app:finite_to_kernel}

We briefly show how the CFKS representer form follows from the finite RBF
feature model. Consider
\begin{equation}
	u_M(\mathbf z)
	=
	\frac{1}{\sqrt{M}}
	\sum_{m=1}^{M}
	\beta_m\phi_{\theta_m}(\mathbf z),
	\label{eq:finite_feature_model_app}
\end{equation}
and let $\{\ell_i\}_{i=1}^{N}$ denote the selected PDE, boundary, and initial
functionals. Define the finite feature matrix
\begin{equation}
	Z_{im}
	=
	\frac{1}{\sqrt{M}}\,
	\ell_i[\phi_{\theta_m}],
\end{equation}
and the feature vector
\begin{equation}
	\boldsymbol{\varphi}_M(\mathbf z)
	=
	\frac{1}{\sqrt{M}}
	\begin{bmatrix}
		\phi_{\theta_1}(\mathbf z) &
		\cdots &
		\phi_{\theta_M}(\mathbf z)
	\end{bmatrix}^{\top}.
\end{equation}
Then
\begin{equation}
	u_M(\mathbf z)
	=
	\boldsymbol{\varphi}_M(\mathbf z)^{\top}\boldsymbol{\beta}.
\end{equation}

For a minimum-norm feature-space solution satisfying the discrete
constraints, the coefficient vector lies in the range of $Z^{\top}$ and may
therefore be written as
\begin{equation}
	\boldsymbol{\beta}
	=
	Z^{\top}\boldsymbol{\alpha},
\end{equation}
for some coefficient vector
$\boldsymbol{\alpha}\in\mathbb{R}^{N}$. Substituting into the finite feature
representation gives
\begin{align}
	u_M(\mathbf z)
	&=
	\boldsymbol{\varphi}_M(\mathbf z)^{\top}
	Z^{\top}\boldsymbol{\alpha}
	\nonumber\\
	&=
	\sum_{j=1}^{N}
	\alpha_j
	\left[
	\frac{1}{M}
	\sum_{m=1}^{M}
	\phi_{\theta_m}(\mathbf z)
	\ell_j[\phi_{\theta_m}]
	\right].
	\label{eq:finite_representer_app}
\end{align}
Because $\ell_j$ acts on the second argument, the term in brackets can be
written as
\begin{equation}
	\ell_j^{\mathbf z'}
	k_M(\mathbf z,\mathbf z'_j),
\end{equation}
where the empirical feature kernel is
\begin{equation}
	k_M(\mathbf z,\mathbf z')
	=
	\frac{1}{M}
	\sum_{m=1}^{M}
	\phi_{\theta_m}(\mathbf z)
	\phi_{\theta_m}(\mathbf z').
\end{equation}
Hence
\begin{equation}
	u_M(\mathbf z)
	=
	\sum_{j=1}^{N}
	\alpha_j
	\ell_j^{\mathbf z'}
	k_M(\mathbf z,\mathbf z'_j).
	\label{eq:finite_kernel_representer_app}
\end{equation}

If the feature parameters are sampled from $\rho(\theta)$, then as
$M\rightarrow\infty$,
\begin{equation}
	k_M(\mathbf z,\mathbf z')
	\longrightarrow
	k_{\rho}(\mathbf z,\mathbf z')
	=
	\mathbb{E}_{\theta\sim\rho}
	\left[
	\phi_{\theta}(\mathbf z)
	\phi_{\theta}(\mathbf z')
	\right].
\end{equation}
Under the regularity conditions used in this work, the linear functionals may
be interchanged with the expectation, so that
\begin{equation}
	\ell_j^{\mathbf z'}
	k_M(\mathbf z,\mathbf z'_j)
	\longrightarrow
	\ell_j^{\mathbf z'}
	k_{\rho}(\mathbf z,\mathbf z'_j).
\end{equation}
The continuum-limit representation is therefore
\begin{equation}
	u_h(\mathbf z)
	=
	\sum_{j=1}^{N}
	\alpha_j
	\ell_j^{\mathbf z'}
	k_{\rho}(\mathbf z,\mathbf z'_j),
	\label{eq:kernel_representer_app}
\end{equation}
which is the CFKS representer form used in the main text.

This derivation also shows that the finite feature coefficients
$\boldsymbol{\beta}$ disappear from the continuum formulation: their effect
is transferred to the constraint-space coefficients
$\boldsymbol{\alpha}$ through the empirical Gram structure, while the sampled
kernel $k_M$ is replaced by the analytically evaluated continuum-feature mean
kernel $k_{\rho}$.

\section{Analytical Derivation of the Bounded Continuum-Feature Gaussian Kernel}
\label{app:bounded_gaussian_kernel}

This appendix derives the closed-form mean kernel associated with a
continuum of Gaussian radial-basis-function features whose centers are
uniformly distributed over a bounded interval. The full-line version of
this construction is classical and yields a squared-exponential
covariance; see, for example, Rasmussen and Williams
\citep{rasmussen2006gaussian}. Here we retain the finite center-support
interval explicitly, which produces additional error-function terms.

\subsection{One-dimensional Gaussian feature model}
\label{app:one_dimensional_feature}

Consider the Gaussian feature
\begin{equation}
	\phi_c(x)
	=
	\exp\left(
	-\frac{(x-c)^2}{2\sigma^2}
	\right),
	\qquad
	\sigma>0,
	\label{eq:app_gaussian_feature}
\end{equation}
where the center $c$ is distributed uniformly over
\begin{equation}
	c\sim\mathcal{U}[a,b],
	\qquad
	a<b.
\end{equation}
The corresponding mean feature kernel is
\begin{equation}
	k_{[a,b]}(x,y)
	=
	\mathbb{E}_{c}
	\left[
	\phi_c(x)\phi_c(y)
	\right]
	=
	\frac{1}{b-a}
	\int_a^b
	\phi_c(x)\phi_c(y)\,dc.
	\label{eq:app_population_kernel_definition}
\end{equation}

Substituting eq. \ref{eq:app_gaussian_feature} gives
\begin{equation}
	k_{[a,b]}(x,y)
	=
	\frac{1}{b-a}
	\int_a^b
	\exp\left(
	-\frac{(x-c)^2+(y-c)^2}{2\sigma^2}
	\right)
	dc.
	\label{eq:app_kernel_before_square}
\end{equation}

The integral can be evaluated analytically by completing the square with
respect to the center variable $c$.

\subsection{Completion of the square}
\label{app:complete_square}

Define the midpoint and separation coordinates
\begin{equation}
	m
	=
	\frac{x+y}{2},
	\qquad
	d
	=
	x-y.
	\label{eq:app_midpoint_separation}
\end{equation}
We first rewrite the quadratic form in the exponent:
\begin{align}
	(x-c)^2+(y-c)^2
	&=
	x^2+y^2-2c(x+y)+2c^2.
	\label{eq:app_expand_quadratic}
\end{align}
Using
\begin{equation}
	x^2+y^2
	=
	\frac{(x-y)^2}{2}
	+
	\frac{(x+y)^2}{2},
\end{equation}
and
\begin{equation}
	2(c-m)^2
	=
	2c^2
	-
	2c(x+y)
	+
	\frac{(x+y)^2}{2},
\end{equation}
we obtain
\begin{equation}
	(x-c)^2+(y-c)^2
	=
	2(c-m)^2
	+
	\frac{(x-y)^2}{2}.
	\label{eq:app_completed_square}
\end{equation}

Substituting eq. \ref{eq:app_completed_square} into
eq. \ref{eq:app_kernel_before_square} gives
\begin{align}
	k_{[a,b]}(x,y)
	&=
	\frac{1}{b-a}
	\int_a^b
	\exp\left(
	-\frac{(x-y)^2}{4\sigma^2}
	\right)
	\exp\left(
	-\frac{(c-m)^2}{\sigma^2}
	\right)
	dc
	\nonumber\\
	&=
	\frac{
		\exp\left(
		-\frac{d^2}{4\sigma^2}
		\right)
	}{b-a}
	\int_a^b
	\exp\left(
	-\frac{(c-m)^2}{\sigma^2}
	\right)
	dc.
	\label{eq:app_separated_kernel}
\end{align}

Thus, the kernel separates into a Gaussian factor depending on the
relative distance $d=x-y$ and a finite-support correction depending on the
midpoint $m=(x+y)/2$.

\subsection{Evaluation using the error function}
\label{app:error_function_evaluation}

Introduce the change of variables
\begin{equation}
	z
	=
	\frac{c-m}{\sigma},
	\qquad
	dc
	=
	\sigma\,dz.
\end{equation}
The limits become
\begin{equation}
	z_a
	=
	\frac{a-m}{\sigma},
	\qquad
	z_b
	=
	\frac{b-m}{\sigma}.
\end{equation}
Therefore,
\begin{align}
	\int_a^b
	\exp\left(
	-\frac{(c-m)^2}{\sigma^2}
	\right)
	dc
	&=
	\sigma
	\int_{z_a}^{z_b}
	e^{-z^2}\,dz.
	\label{eq:app_gaussian_integral}
\end{align}

Using the error function
\begin{equation}
	\operatorname{erf}(z)
	=
	\frac{2}{\sqrt{\pi}}
	\int_0^z
	e^{-t^2}\,dt,
\end{equation}
we have
\begin{equation}
	\int_{z_a}^{z_b}
	e^{-z^2}\,dz
	=
	\frac{\sqrt{\pi}}{2}
	\left[
	\operatorname{erf}(z_b)
	-
	\operatorname{erf}(z_a)
	\right].
\end{equation}
Hence,
\begin{align}
	\int_a^b
	\exp\left(
	-\frac{(c-m)^2}{\sigma^2}
	\right)
	dc
	=
	\frac{\sqrt{\pi}\sigma}{2}
	\left[
	\operatorname{erf}
	\left(
	\frac{b-m}{\sigma}
	\right)
	-
	\operatorname{erf}
	\left(
	\frac{a-m}{\sigma}
	\right)
	\right].
	\label{eq:app_finite_gaussian_integral}
\end{align}

Substitution into eq. \ref{eq:app_separated_kernel} yields the closed form
\begin{equation}
	\boxed{
		\begin{aligned}
			k_{[a,b]}(x,y)
			&=
			\frac{\sqrt{\pi}\sigma}{2(b-a)}
			\exp\left(
			-\frac{(x-y)^2}{4\sigma^2}
			\right)
			\\
			&\quad\times
			\left[
			\operatorname{erf}
			\left(
			\frac{
				b-\frac{x+y}{2}
			}{\sigma}
			\right)
			-
			\operatorname{erf}
			\left(
			\frac{
				a-\frac{x+y}{2}
			}{\sigma}
			\right)
			\right].
		\end{aligned}
	}
	\label{eq:app_bounded_gaussian_kernel}
\end{equation}

Equation eq. \ref{eq:app_bounded_gaussian_kernel} is the bounded
continuum-feature Gaussian kernel used in this work.

\subsection{Interpretation of the bounded-support correction}
\label{app:bounded_interpretation}

The expression in eq. \ref{eq:app_bounded_gaussian_kernel} can be written
schematically as
\begin{equation}
	k_{[a,b]}(x,y)
	=
	k_{\mathrm{rel}}(x-y)
	k_{\mathrm{sup}}
	\left(
	\frac{x+y}{2}
	\right),
\end{equation}
where
\begin{equation}
	k_{\mathrm{rel}}(x-y)
	=
	\exp\left(
	-\frac{(x-y)^2}{4\sigma^2}
	\right)
\end{equation}
depends only on the point separation, while
\begin{equation}
	\begin{aligned}
		k_{\mathrm{sup}}(m)
		&=
		\frac{\sqrt{\pi}\sigma}{2(b-a)}
		\left[
		\operatorname{erf}
		\left(
		\frac{b-m}{\sigma}
		\right)
		-
		\operatorname{erf}
		\left(
		\frac{a-m}{\sigma}
		\right)
		\right]
	\end{aligned}
\end{equation}
depends on the midpoint $m=(x+y)/2$.

This decomposition shows explicitly why the bounded-center kernel is
generally nonstationary. If the center support is finite, two pairs of
points with the same separation $x-y$ can have different covariance if
their midpoints lie at different distances from the center-support
boundaries.

Away from the edges $a$ and $b$, and provided the support is wide relative
to $\sigma$, the two error functions approach $+1$ and $-1$,
respectively, so that the support correction becomes approximately
constant. Near the edges, however, the finite-support correction modifies
the kernel.

\subsection{Full-line limit}
\label{app:full_line_limit}

The relationship with the standard Gaussian or squared-exponential
kernel is obtained by taking
\begin{equation}
	a\rightarrow-\infty,
	\qquad
	b\rightarrow+\infty.
\end{equation}
Since
\begin{equation}
	\lim_{z\rightarrow+\infty}\operatorname{erf}(z)=1,
	\qquad
	\lim_{z\rightarrow-\infty}\operatorname{erf}(z)=-1,
\end{equation}
the bracketed factor in
eq. \ref{eq:app_bounded_gaussian_kernel} tends to $2$.

If one considers the unnormalized center measure
\begin{equation}
	d\mu(c)=dc,
\end{equation}
rather than the uniform probability density $1/(b-a)$, the corresponding
full-line feature integral is
\begin{align}
	\widetilde{k}_{\mathbb{R}}(x,y)
	&=
	\int_{-\infty}^{\infty}
	\phi_c(x)\phi_c(y)\,dc
	\nonumber\\
	&=
	\sqrt{\pi}\sigma
	\exp\left(
	-\frac{(x-y)^2}{4\sigma^2}
	\right).
	\label{eq:app_full_line_kernel}
\end{align}

After normalization of the amplitude, this is a squared-exponential
kernel with effective length scale
\begin{equation}
	\ell
	=
	\sqrt{2}\,\sigma,
\end{equation}
under the conventional form
\begin{equation}
	k_{\mathrm{SE}}(x,y)
	=
	\exp\left(
	-\frac{(x-y)^2}{2\ell^2}
	\right).
\end{equation}

Thus, the usual stationary Gaussian kernel is recovered as the
infinite-support limit of the Gaussian center-feature construction, while
the bounded-center model retains additional support information.

\subsection{Positive semidefiniteness}
\label{app:positive_semidefinite}

The positive semidefiniteness of
eq. \ref{eq:app_population_kernel_definition} follows immediately from its
feature representation. For arbitrary points
$\{x_i\}_{i=1}^{N}$ and coefficients
$\{\alpha_i\}_{i=1}^{N}$,
\begin{align}
	\sum_{i=1}^{N}
	\sum_{j=1}^{N}
	\alpha_i\alpha_j
	k_{[a,b]}(x_i,x_j)
	&=
	\mathbb{E}_{c}
	\left[
	\sum_{i,j}
	\alpha_i\alpha_j
	\phi_c(x_i)\phi_c(x_j)
	\right]
	\nonumber\\
	&=
	\mathbb{E}_{c}
	\left[
	\left(
	\sum_{i=1}^{N}
	\alpha_i\phi_c(x_i)
	\right)^2
	\right]
	\geq 0.
	\label{eq:app_psd_proof}
\end{align}

Hence, the continuum-center construction defines a valid positive
semidefinite kernel independently of the closed-form calculation.

\subsection{Differentiation of the mean kernel}
\label{app:kernel_differentiation}

For PDE applications, the essential quantity is not only
$k_{[a,b]}(x,y)$ itself but its derivatives with respect to both arguments.
Since the Gaussian feature and all of its derivatives are smooth, and the
center support $[a,b]$ is bounded, differentiation may be interchanged
with integration:
\begin{equation}
	\partial_x^p
	\partial_y^q
	k_{[a,b]}(x,y)
	=
	\frac{1}{b-a}
	\int_a^b
	\partial_x^p\phi_c(x)
	\,
	\partial_y^q\phi_c(y)
	\,dc.
	\label{eq:app_derivative_expectation}
\end{equation}

Equation \ref{eq:app_derivative_expectation} gives a direct mean
interpretation of every differentiated kernel block.

For example,
\begin{equation}
	\partial_x\phi_c(x)
	=
	-\frac{x-c}{\sigma^2}
	\phi_c(x),
\end{equation}
and
\begin{equation}
	\partial_x^2\phi_c(x)
	=
	\left[
	\frac{(x-c)^2}{\sigma^4}
	-
	\frac{1}{\sigma^2}
	\right]
	\phi_c(x).
\end{equation}

Therefore,
\begin{equation}
	\partial_x\partial_y
	k_{[a,b]}(x,y)
	=
	\mathbb{E}_{c}
	\left[
	\partial_x\phi_c(x)
	\partial_y\phi_c(y)
	\right],
\end{equation}
and similarly,
\begin{equation}
	\partial_x^2
	\partial_y^2
	k_{[a,b]}(x,y)
	=
	\mathbb{E}_{c}
	\left[
	\partial_x^2\phi_c(x)
	\partial_y^2\phi_c(y)
	\right].
\end{equation}

For a differential operator $\mathcal{L}$,
\begin{equation}
	\boxed{
		\mathcal{L}_x
		\mathcal{L}_y
		k_{[a,b]}(x,y)
		=
		\mathbb{E}_{c}
		\left[
		(\mathcal{L}\phi_c)(x)
		(\mathcal{L}\phi_c)(y)
		\right].
	}
	\label{eq:app_operator_population_identity}
\end{equation}

Thus, the operator kernel used in the continuum-feature collocation method
is precisely the mean Gram kernel of the operator-applied Gaussian
features.

\subsection{Gaussian moment representation}
\label{app:gaussian_moments}

Higher-order derivatives can be evaluated efficiently without repeatedly
differentiating the error-function expression
eq. \ref{eq:app_bounded_gaussian_kernel}. After completing the square, every
mixed derivative reduces to finite combinations of Gaussian moments of
the form
\begin{equation}
	J_n(m)
	=
	\int_a^b
	(c-m)^n
	\exp\left(
	-\frac{(c-m)^2}{\sigma^2}
	\right)
	dc.
	\label{eq:app_gaussian_moments}
\end{equation}

The first moments are
\begin{equation}
	J_0
	=
	\frac{\sqrt{\pi}\sigma}{2}
	\left[
	\operatorname{erf}
	\left(
	\frac{b-m}{\sigma}
	\right)
	-
	\operatorname{erf}
	\left(
	\frac{a-m}{\sigma}
	\right)
	\right],
\end{equation}
and
\begin{equation}
	J_1
	=
	\frac{\sigma^2}{2}
	\left[
	\exp\left(
	-\frac{(a-m)^2}{\sigma^2}
	\right)
	-
	\exp\left(
	-\frac{(b-m)^2}{\sigma^2}
	\right)
	\right].
\end{equation}

For $n\geq2$, integration by parts gives the recursion
\begin{equation}
	\boxed{
		J_n
		=
		-\frac{\sigma^2}{2}
		\left[
		(b-m)^{n-1}
		e^{-(b-m)^2/\sigma^2}
		-
		(a-m)^{n-1}
		e^{-(a-m)^2/\sigma^2}
		\right]
		+
		\frac{(n-1)\sigma^2}{2}
		J_{n-2}.
	}
	\label{eq:app_moment_recursion}
\end{equation}

This recursion provides a convenient analytic route to the high-order
kernel derivatives required by elliptic, advection--diffusion, and
biharmonic operators.

\subsection{Extension to multiple dimensions}
\label{app:multi_dimensional_extension}

Suppose the $d$-dimensional Gaussian feature is separable,
\begin{equation}
	\phi_{\boldsymbol{c}}(\boldsymbol{x})
	=
	\prod_{r=1}^{d}
	\exp\left(
	-\frac{(x_r-c_r)^2}{2\sigma_r^2}
	\right),
	\label{eq:app_multidim_feature}
\end{equation}
and the center coordinates are independently distributed as
\begin{equation}
	c_r
	\sim
	\mathcal{U}[a_r,b_r].
\end{equation}

The mean density factorizes,
\begin{equation}
	\rho(\boldsymbol{c})
	=
	\prod_{r=1}^{d}
	\frac{1}{b_r-a_r},
\end{equation}
and therefore the multidimensional kernel also factorizes:
\begin{align}
	k_{\rho}
	(\boldsymbol{x},\boldsymbol{y})
	&=
	\mathbb{E}_{\boldsymbol{c}}
	\left[
	\phi_{\boldsymbol{c}}(\boldsymbol{x})
	\phi_{\boldsymbol{c}}(\boldsymbol{y})
	\right]
	\nonumber\\
	&=
	\prod_{r=1}^{d}
	k_{[a_r,b_r]}^{(r)}
	(x_r,y_r).
	\label{eq:app_tensor_product_kernel}
\end{align}

Consequently, mixed derivatives also factorize:
\begin{equation}
	D_{\boldsymbol{x}}^{\boldsymbol{\alpha}}
	D_{\boldsymbol{y}}^{\boldsymbol{\beta}}
	k_{\rho}
	(\boldsymbol{x},\boldsymbol{y})
	=
	\prod_{r=1}^{d}
	\partial_{x_r}^{\alpha_r}
	\partial_{y_r}^{\beta_r}
	k_{[a_r,b_r]}^{(r)}
	(x_r,y_r).
	\label{eq:app_tensor_derivatives}
\end{equation}

This property is particularly useful for Cartesian embedding domains and
space--time formulations. For example, a two-dimensional Laplacian kernel
can be assembled from one-dimensional derivative factors, while
space--time advection--diffusion kernels can be constructed from separate
spatial and temporal mean kernels.

\subsection{Summary}
\label{app:derivation_summary}

The continuum-center Gaussian feature model therefore admits the exact
bounded-support kernel
\begin{equation}
	k_{[a,b]}(x,y)
	=
	\frac{\sqrt{\pi}\sigma}{2(b-a)}
	e^{-(x-y)^2/(4\sigma^2)}
	\left[
	\operatorname{erf}
	\left(
	\frac{b-\frac{x+y}{2}}{\sigma}
	\right)
	-
	\operatorname{erf}
	\left(
	\frac{a-\frac{x+y}{2}}{\sigma}
	\right)
	\right],
\end{equation}
which reduces to a Gaussian/squared-exponential kernel in the full-support
limit after normalization.

More importantly for PDE collocation, the differentiated mean
kernel satisfies
\begin{equation}
	D_x^\alpha D_y^\beta k_\rho(x,y)
	=
	\mathbb{E}_{c\sim\rho}
	\left[
	D_x^\alpha\phi_c(x)
	D_y^\beta\phi_c(y)
	\right].
\end{equation}
Hence, the differential-operator kernel matrices used by the proposed
method are exactly the continuum limits of the corresponding finite
Gaussian-RBF operator Gram matrices.

\section*{CRediT authorship contribution statement}
\textbf{Vikas Dwivedi:} Conceptualization, Methodology, Software, Writing - Original Draft, 
\textbf{Bruno Sixou} and \textbf{Monica Sigovan:} Supervision and Writing - Review \& Editing.

\section*{Acknowledgements}
This work was supported by the ANR (Agence Nationale de la Recherche), France, through the RAPIDFLOW project (Grant no. ANR-24-CE19-1349-01). 

\bibliographystyle{unsrt}  
\bibliography{references}

\end{document}